\documentclass{article}
\usepackage{ijcai25}

\usepackage{times}
\usepackage{soul}
\usepackage{url}
\usepackage{xcolor}
\definecolor{MyCiteColor}{rgb}{0, 0.6, 1}
\usepackage[colorlinks,citecolor=MyCiteColor]{hyperref}
\usepackage[utf8]{inputenc}
\usepackage[small]{caption}
\usepackage{graphicx}
\usepackage{amsmath}
\usepackage{amsthm}
\usepackage{booktabs}
\usepackage{algorithm}
\usepackage[switch]{lineno}
\usepackage{algorithmic}
\usepackage{colortbl}

\usepackage{makecell,multirow,diagbox}
\usepackage{color}
\usepackage{mathtools,xspace}
\usepackage{subcaption}
\usepackage{bbm}
\usepackage{amssymb}
\usepackage{dsfont}
\usepackage{bbding}
\usepackage{pifont}
\usepackage{hyperref}
\usepackage[T1]{fontenc}

\newcommand{\onedot}{\ifx\@let@token.\else.\null\fi\xspace}
\newcommand{\etal}{\emph{et al}\onedot}

\newcommand{\ie}{\emph{i.e}\onedot}
\newcommand{\etc}{\emph{etc}}

\newcommand{\darkgreen}[1]{\textcolor[rgb]{0.00,0.70,0.00}{#1}}
\newcommand{\darkred}[1]{\textcolor[rgb]{0.70,0.00,0.00}{#1}}
\newcommand{\darkblue}[1]{\textcolor[rgb]{0.00,0.00,0.90}{#1}}

\title{

Reasoning-OCR: Can Large Multimodal Models Solve Complex Logical Reasoning Problems from OCR Cues?
}

\author{
Haibin He$^1$
\and
Maoyuan Ye$^1$\and
Jing Zhang$^1$\and
Xiantao Cai$^1$\and
Juhua Liu$^1$\and
Bo Du$^1$\And
Dacheng Tao$^2$
\affiliations
$^1$School of Computer Science, National Engineering Research Center for Multimedia Software, and Institute of Artificial Intelligence, Wuhan University, China\\
$^2$College of Computing \& Data Science at Nanyang Technological University\\
\emails
\{haibinhe, yemaoyuan, caixiantao, liujuhua, dubo\}@whu.edu.cn,
jingzhang.cv@gmail.com,
dacheng.tao@gmail.com
}

\begin{document}

\maketitle

\begin{abstract}
Large Multimodal Models (LMMs) have become increasingly versatile, accompanied by impressive Optical Character Recognition (OCR) related capabilities. 
Existing OCR-related benchmarks emphasize evaluating LMMs' abilities of relatively simple visual question answering, visual-text parsing, \etc. However, the extent to which LMMs can deal with complex logical reasoning problems based on OCR cues is relatively unexplored.  
To this end, we introduce the \textbf{Reasoning-OCR} benchmark, which challenges LMMs to solve complex reasoning problems based on the cues that can be extracted from rich visual-text. 
Reasoning-OCR covers six visual scenarios and encompasses 150 meticulously designed questions categorized into six reasoning challenges. Additionally, Reasoning-OCR minimizes the impact of field-specialized knowledge. 
Our evaluation offers some insights for proprietary and open-source LMMs in different reasoning challenges, underscoring the urgent to improve the reasoning performance.
We hope Reasoning-OCR can inspire and facilitate future research on enhancing complex reasoning ability based on OCR cues. Reasoning-OCR is publicly available at \url{https://github.com/Hxyz-123/ReasoningOCR}.
\end{abstract}

\section{Introduction}
Visual-text in images contains essential information for multimodal intelligence. Given text-rich input images, the OCR ability has become the foundation of LMMs~\cite{achiam2023gpt,li2024llava,wang2024qwen2,chen2024internvl} to further interact with humans in depth. 
Thanks to the large-scale and diverse training data, recent LMMs deliver impressive OCR-related performance, such as text recognition~\cite{liu2024ocrbench} and document parsing~\cite{wei2024general}. Nonetheless, merely reading and understanding visual-text cannot enable LMMs to successfully solve more complicated tasks like planning and decision making (such as for embodied artificial intelligence), underlining the importance of developing and evaluating complex logical reasoning ability based on OCR cues.

To assess visual-text-based capabilities of LMMs, most existing benchmarks~\cite{liu2024ocrbench,zhang2024exploring,wadhawancontextual,tang2024mtvqa} focus on evaluating the narrowly or broadly defined OCR tasks and relatively simple visual question answering (VQA), rather than complex logical reasoning problems. For example, Liu \etal~\cite{liu2024focus} provide a dataset for benchmarking single- and multi-page document parsing in page level or local region. TextVQA~\cite{singh2019towards} and DocVQA~\cite{mathew2021docvqa} are widely used benchmarks for evaluating scene- and document-orient VQA. However, these benchmarks lack samples that require complex reasoning. Some benchmarks even rapidly reach performance saturation, such as 95.3\% scores of InternVL2.5~\cite{internvl2_5} on DocVQA, probably leading to over-optimistic measure of progress toward advanced intelligence. 
Although some chart benchmarks~\cite{chen2024onechart,xia2024chartx,liu2024mmc,xu2023chartbench} provide reasoning problems, the visual scenario is constrained to the sole chart context.

\begin{figure*}[t]
\centering
\includegraphics[width=\linewidth]{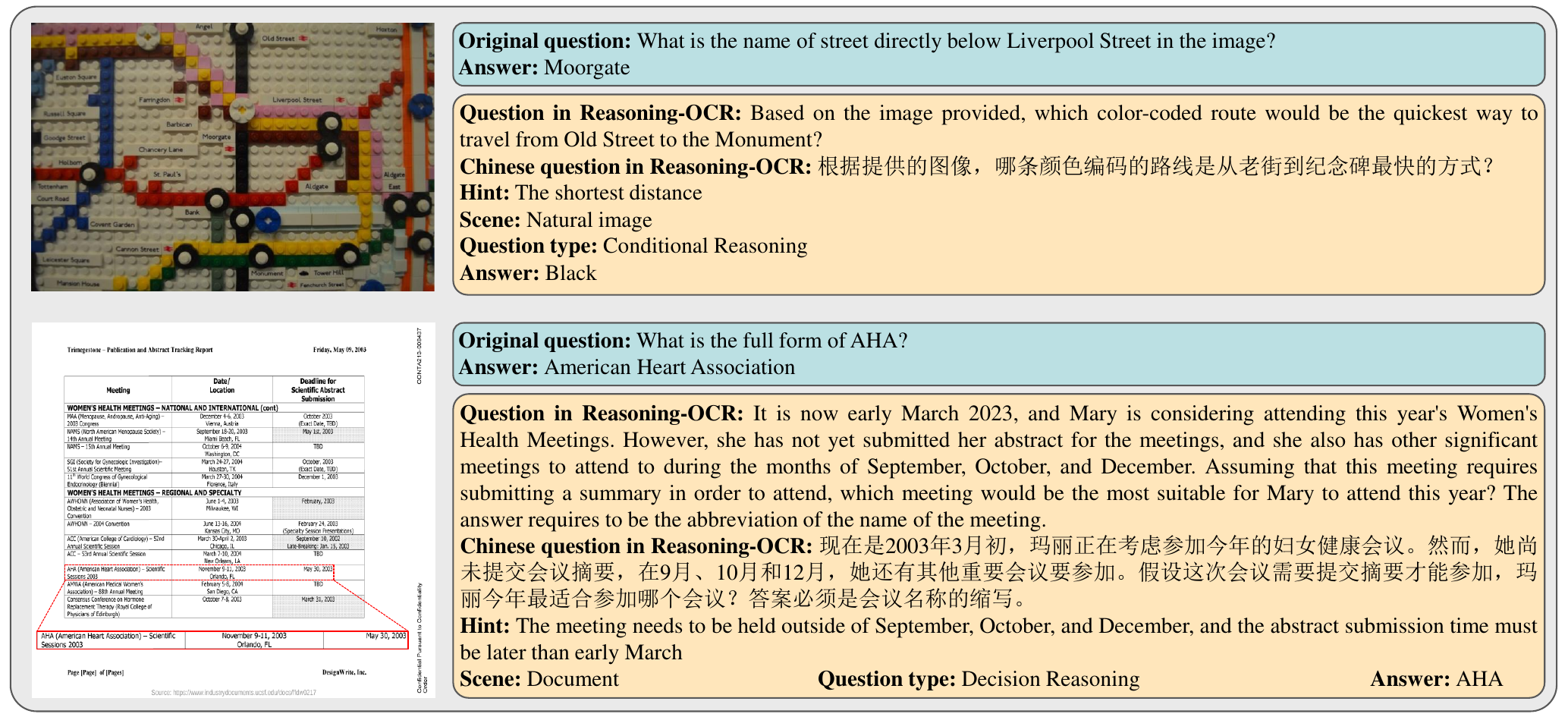} 
\caption{\textbf{Samples in Reasoning-OCR.} Relevant textual cues in the image are highlighted with red circles or rectangles for clarity, which are not visible to LMMs. Our questions demonstrate higher reasoning complexity compared to the ones in source datasets.}
\label{example}
\end{figure*}

To fill the gap, we introduce the \textbf{Reasoning-OCR} benchmark, featuring complex logical reasoning problems that can only be answered with OCR cues and covering diverse visual scenarios (including chart, product label, document, natural image, screen shot, and token). In this benchmark, the reasoning type are categorized into six classes, \ie, 1) \textit{data comparison analysis}, 2) \textit{data statistical analysis}, 3) \textit{mathematical reasoning}, 4) \textit{conditional reasoning}, 5) \textit{temporal reasoning}, and 6) \textit{decision reasoning}. To ensure high complexity and data quality, we filter images from publicly available datasets and web-source, then meticulously design 150 questions. We annotate questions in both English and Chinese, and also provide the key answer hint for each question. Two examples about conditional and decision reasoning are shown in Fig.~\ref{example}.
Different from mathematics benchmarks~\cite{wang2024measuring,zhang2025mathverse,he2024olympiadbench}, we minimize the requirement for prior and field-specialized knowledge.

For comprehensive evaluation, we select 9 open-source LMMs and the proprietary GPT-4o~\cite{openai2024gpt4o}, including both generic and text-centric LMMs. We analyze the LMMs in different reasoning types and explore the influence of chain-of-thought (CoT), using answer hint, \etc. Moreover, we also analysis the error source of LMMs in details. 
Through our evaluation, some key findings are revealed as follows. 
\ding{182} There is huge potential for improving the complex logical reasoning capabilities using OCR cues. 
\ding{183} Text-centric LMMs lag far behind advanced generic LMMs on reasoning. 
\ding{184} LMMs fall short in decision reasoning. 
\ding{185} CoT generally helps LMMs achieve better overall accuracy.

To summarize, our main contributions are three-fold: 
\begin{itemize}
    \item We construct Reasoning-OCR, a challenging benchmark designed for evaluating the complex logical reasoning capabilities using OCR cues for LMMs while minimizing the impact of field-specialized knowledge.
    \item We conduct a comprehensive evaluation on proprietary and open-source LMMs, including both generic and text-centric ones. We provide in-depth analysis on LMMs in diverse reasoning tasks and various settings.
    \item Our provided benchmark and insights could facilitate future research and point out the improvement directions for incoming LMMs.
\end{itemize}

\section{Related Work}
\label{sec: Related Work}
\subsection{Large Multimodal Models}
Taking advantage of multimodal pre-training, early LMMs \cite{dai2023instructblip,alayrac2022flamingo} demonstrate certain OCR capabilities. Nonetheless, their performance is inferior to some specialist models~\cite{liu2024ocrbench}. 
In contrast, recent LMMs~\cite{ye2023ureader,liu2024improved,li2024llava,mplug-docowl2,wei2024general} have obtained favorable OCR performance via improving image processing pipeline and curating training data. 
On the one hand, Monkey~\cite{li2024monkey} and TextMonkey~\cite{liu2024textmonkey} enable processing high resolution images by dividing them into uniform patches, promoting the perception of visual-text. 
InternVL2~\cite{chen2024internvl} deals with input images with a set of pre-defined aspect ratios. Instead of using predetermined aspect ratios, Qwen2-VL~\cite{wang2024qwen2} encodes images based on their original resolution. 
On the other hand, regarding training samples, both InternVL2 and Qwen2-VL inject abundant OCR data. 
As a result, these OCR-enhanced LMMs are competent for text recognition, document elements parsing, text-based VQA, \etc.
However, beyond mere OCR, whether these LMMs can successfully solve complex reasoning problems based on OCR cues is under exploration.

\subsection{OCR-Related Benchmarks for LMMs}
As LMMs are in the ascendant towards advanced literacy, numerous benchmarks~\cite{singh2019towards,mathew2021docvqa,wadhawancontextual,zhang2024exploring,liu2024focus,li2024seed} have emerged to evaluate OCR-related abilities of LMMs in higher fidelity. For instance, OCRBench~\cite{liu2024ocrbench} evaluates LMMs' performance on text and handwritten expression recognition, key information extraction, scene- and document-oriented VQA. CC-OCR~\cite{yang2024cc} leans toward benchmarking document parsing as well as OCR under multi-scene and multi-lingual settings.
Although they comprehensively assess practical OCR and common VQA abilities, they lack evaluating the complex logical reasoning performance with OCR cues.

For evaluating reasoning ability of LMMs, chart and mathematical VQA benchmarks are commonly chosen. Specifically, CharXiv~\cite{wangcharxiv} consists of both descriptive and reasoning questions on diverse chart images, posing more difficulties compared to ChartQA~\cite{masry2022chartqa}. 
GeoQA~\cite{chen2021geoqa} and MathVista~\cite{lumathvista} present challenges for mathematical reasoning in visual contexts. 
Compared to the above benchmarks, our benchmark contains more diverse scenarios, including chart, document, screen shot, tokens, \etc. In addition, compared to mathematical benchmarks, our benchmark focuses on complex reasoning within text-rich images while minimizing the impact of prior and specialized knowledge, such as geometry.

\begin{figure}[t]
\centering
\includegraphics[width=0.75\linewidth]{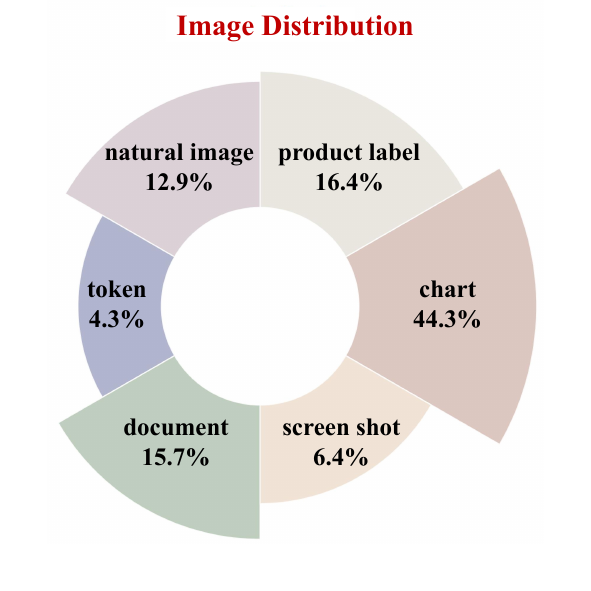} 
\caption{\textbf{The visual scenario distribution in Reasoning-OCR.} The collected images cover six visual scenarios, including chart, product label, document, natural image, screen shot, and token (from the most to the least).}
\label{img_stat}
\end{figure}

\section{Reasoning-OCR Benchmark}
Existing visual-text-based benchmarks~\cite{masry2022chartqa,mathew2021docvqa,zhang2024exploring,liu2024ocrbench} often exhibit limited problem complexity, rendering them insufficient for effectively assessing the multi-hop reasoning capabilities of contemporary LMMs. To better evaluate the complex logical reasoning abilities of LMMs, we collect a novel dataset Reasoning-OCR. It comprises 140 images and 150 meticulously crafted problems based on textual cues from the images, encompassing diverse scenarios and problem types. Two examples from the dataset are illustrated in Fig.~\ref{example}. 

\subsection{Dataset Construction}
\noindent\textbf{Image Collection.} 
Images in our Reasoning-OCR are collected from four parts: 1) 60 images (42.9\%) from ChartQA~\cite{masry2022chartqa}; 2) 50 images (35.7\%) from DT-VQA~\cite{zhang2024exploring}; 3) 20 images (14.3\%) from DocVQA~\cite{mathew2021docvqa}; and 4) the rest 10 images (7.1\%) are obtained from the web. 
These images encompass a variety of scenarios and can be further categorized into six distinct groups: charts, product labels, documents (including printed documents, electronic documents, \etc), natural images, screenshots (such as excerpts from medical books, segments of product instructions and specifications), and tokens (including paper currency and tickets). 

\noindent\textbf{Question Design.} Different from previous OCR-related benchmarks, the questions in Reasoning-OCR are thoughtfully designed by experts based on the textual cues from images, featuring more complex reasoning and significantly longer question length. For ease of evaluation, all questions are formulated as objective-type questions, with answers that are both concise and unambiguous. On average, each question requires approximately \textit{half an hour}, encompassing the selection of appropriate images, as well as the question design and annotation. In total, this amounts to the equivalent of about two weeks of full working days. Notably, certain reasoning tasks are highly applicable to the domain of embodied agents, particularly in areas such as conditional reasoning and decision reasoning. This highlights the benchmark's versatility and its potential to drive advancements in these critical areas of research and application.

\begin{figure*}[t]
\centering
\includegraphics[width=\linewidth]{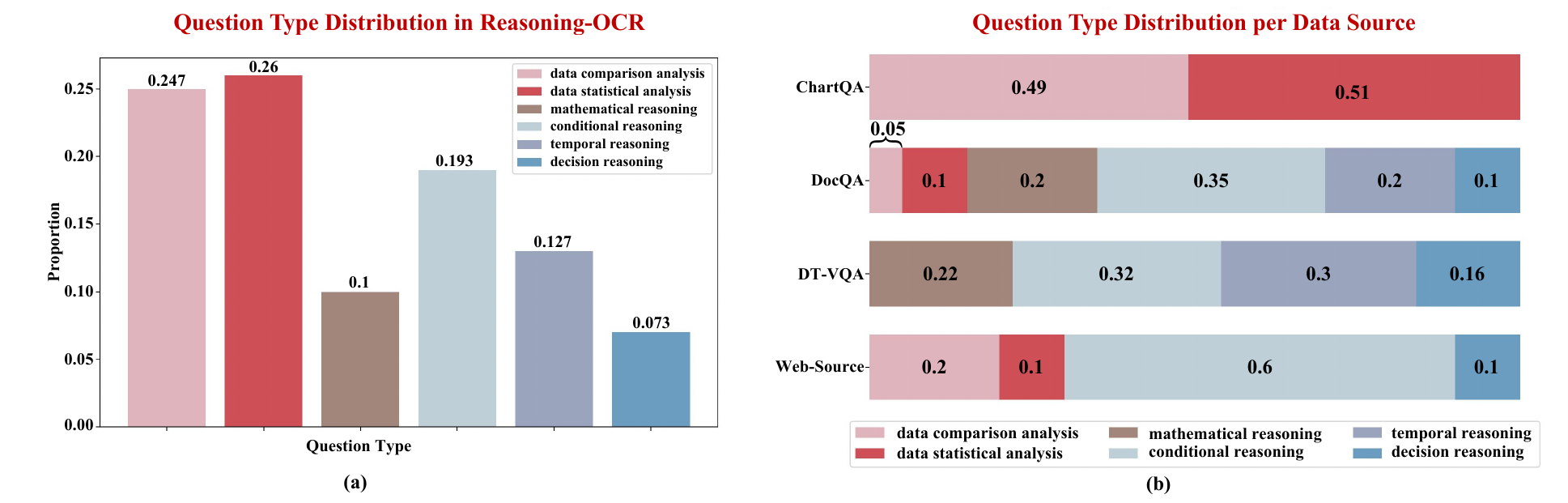} 
\caption{\textbf{The distribution of question types.} (a) shows the distribution of the six question types across the Reasoning-OCR while (b) describes the distribution of the six question types within each of the data sources.}
\label{question_stat}
\end{figure*}

\noindent\textbf{Annotation.} To record the details of each QA, the annotations contain the following dictionary keys: 1) \textit{img}: the image name; 2) \textit{q\_id}: the question index; 3) \textit{question}: the content of the question; 4) \textit{question\_c}: the Chinese version of the question; 5) \textit{answer}: the concise answer to the question; 
6) \textit{hint}: a hint provided to aid in answering the question; 7) \textit{datasource}: the source of the image; 8) \textit{scene}: the scene category associated with the image; 9) \textit{type}: the reasoning type of the question, which includes six categories as shown in Fig.~\ref{question_stat}.

\subsection{Statistics and Characteristics}
Our Reasoning-OCR consists of 150 elaborately designed questions framed on 140 images. It contains six primary scenarios, with the detailed distribution across these scenarios presented in Fig.~\ref{img_stat}. As previously mentioned, the questions are categorized into six distinct types, and their distributions are illustrated in Fig.~\ref{question_stat}. Each question type represents a specific aspect of reasoning, ensuring wide coverage of diverse logical challenges. For example, \textit{data comparison analysis} and \textit{data statistical analysis} refer to comparing distinct data groups within the image based on the question at hand and performing statistical analysis on a single data group, respectively. Examples of each question type from Reasoning-OCR are provided in the supplementary materials.

In the following, we will present a detailed comparison highlighting the major distinctions between our benchmark and previous OCR-related benchmarks: 

\noindent\textbf{Higher Complexity in Questions.} In addition to the OCR and information extraction capabilities, which are the primary focus of previous benchmarks, our benchmark imposes stricter demands on the LMMs' ability to comprehend complex and unconventional problems and perform multi-hops reasoning, as showcased in the example presented in Fig.~\ref{example}. 

\noindent\textbf{Cross-Linguistic Reasoning Evaluation.} Expect the questions in English, we also prepare Chinese version of the questions as shown in Fig.~\ref{example}. We require the LMMs to answer the questions in Chinese using English in order to assess their cross-linguistic reasoning capabilities.

\begin{table}[t]
\centering
\setlength{\tabcolsep}{6pt}
\resizebox{0.9\columnwidth}{!}{\begin{tabular}{c|c|c|c|c}
\toprule[1pt]
 & ChartQA & Doc-VQA & DT-VQA & Ours\\
\midrule[0.5pt]
Average & 14.2 & 11.4 & 11.9 & \textbf{46.1}  \\
Max & 40 & 27 & 36 & \textbf{276} \\
\bottomrule[1pt]
\end{tabular}}
\caption{\textbf{Comparison of question token counts with related datasets.} Our benchmark contains significantly longer questions, indicating the higher complexity for LMMs.}
\label{tab:token}
\end{table}

\begin{table*}[t]
\centering
\footnotesize
\setlength{\tabcolsep}{2pt}
\resizebox{1\linewidth}{!}{\begin{tabular}{l|cc|cc|cc|cc|cc|cc|cc}
\toprule[1pt]
\multirow{2}{*}{Methods} &\multicolumn{2}{c|}{Data$_c$} &\multicolumn{2}{c|}{Data$_s$} &\multicolumn{2}{c|}{Reasoning$_m$} &\multicolumn{2}{c|}{Reasoning$_c$} &\multicolumn{2}{c|}{Reasoning$_t$} &\multicolumn{2}{c|}{Reasoning$_d$} &\multicolumn{2}{c}{Average}\\ \cline{2-15}
& ACC & ACC$_t$ & ACC & ACC$_t$ & ACC & ACC$_t$ & ACC & ACC$_t$ & ACC & ACC$_t$ & ACC & ACC$_t$ & ACC & ACC$_t$\\ 
\hline
\rowcolor{gray!15} \multicolumn{15}{c}{\textit{Open-source LMMs}} \\
TextMonkey~\cite{liu2024textmonkey} $\diamond$ & 21.6 & 18.9 & 5.1 & 7.7 & 13.3 & 13.3 & 65.5 & 65.5 & 10.5 & 10.5 & 18.2 & 18.2 & 22.4 & 22.4 \\
mPLUG-DocOwl2~\cite{mplug-docowl2} $\diamond$ & 16.2 & 21.6 & 30.8 & 28.2 & 0.0 & 0.0 & 55.2 & 55.2 & 42.1 & 36.8 & 0.0 & 9.1 & 24.1 & 25.2 \\
LLaVA-Next-7B~\cite{llavaNext} & 16.2 & 13.5 & 20.5 & 17.9 & 6.7 & 20.0 & 37.9 & 55.2 & 26.3 &15.8 & 0.0 & 0.0 & 17.9 & 20.4  \\
LLaVA-Next-13B~\cite{llavaNext} & 24.3 & 32.4 & 17.9 & 15.4 & 6.7 & 6.7 & 41.4 & 44.8 & 21.1 & 26.3 & 9.1 & 9.1 & 20.1 & 22.5 \\
LLaVA-OV-7B~\cite{llavaOV} & 27.0 & 27.0 & 30.8 & 38.5 & 13.3 & 20.0 & 65.5 & 62.1 & 42.1 & 36.8 & 9.1 & 0.0 & 31.3 & 30.7 \\
Qwen2-VL-7B~\cite{wang2024qwen2} & 37.8 & 35.1 & 30.8 & 38.5 & 40.0 & 46.7 & 65.5 & 58.6 & 36.8 & 42.1 & 27.3 & 9.1 & 39.7 & 38.4 \\
InternVL2.5-8B~\cite{internvl2_5} & 64.9 & 67.6 & 41.0 & 41.0 & 53.3 & 60.0 & 55.2 & 69.0 & 63.2 & 57.9 & 9.1 & 36.4 & 47.8 & 55.3 \\
InternVL2.5-26B~\cite{internvl2_5} & 67.6 & 62.2 & 61.5 & 61.5 & 46.7 & 46.7 & 62.1 & 72.4 & 47.4 & 57.9 & 36.4 & 36.4 & 53.6 & 56.2 \\
InternVL2.5-38B~\cite{internvl2_5} & 75.7 & 81.1 & 61.5 & 64.1 & 66.7 & 60.0 & 69.0 & 75.9 & 68.4 & 57.9 & 36.4 & 18.2 & 63.0 & 59.5 \\
\rowcolor{gray!15} \multicolumn{15}{c}{\textit{Proprietary LMMs}} \\
GPT-4o-20240806~\cite{openai2024gpt4o} & 78.4 & 70.3 & 79.5 & 76.9 & 73.3 & 86.7 & 72.4 & 75.9 & 68.4 & 68.4 & 36.4 & 18.2 & 68.1 & 66.1 \\
\hline
\hline
Average & 43.0 & 43.0 & 37.9 & 39.0 & 32.0 & 36.0 & 59.0 & 63.5 & 42.6 & 41.0 & 18.2 & 15.3 & 38.8 & 39.7 \\
\bottomrule[1pt]
\end{tabular}}
\caption{\textbf{Performance comparison of state-of-the-art LMMs on Reasoning-OCR.} `$\diamond$' represents text-centric LMMs. In addition to the overall performance, the results for each reasoning type are presented separately. Data$_c$ and Data$_s$ denote data comparison and data statistical analysis, respectively. Reasoning$_m$, Reasoning$_c$, Reasoning$_t$, and Reasoning$_d$ represent mathematical, conditional, temporal, and decision reasoning. \textit{ACC} and \textit{ACC$_t$} denote the outcomes achieved with Chain-of-Thought (CoT) and task-specific instruction, respectively.}
\label{tab:main}
\end{table*}

\noindent\textbf{More Question Tokens.} As illustrated in Tab.~\ref{tab:token}, we employ the tokenizer from InternVL2.5~\cite{internvl2_5} to process the questions into tokens and compute the average and maximum number of tokens per question across the benchmarks. Compared to previous benchmarks, our questions exhibit a significantly greater number of tokens, with both the average and maximum token counts exceeding those of earlier benchmarks—more than three times the average and nearly seven times the maximum.

\begin{table}[t]
\centering
\setlength{\tabcolsep}{2pt}
\resizebox{\columnwidth}{!}{\begin{tabular}{l|c|c|c|c|c}
\toprule[1pt]
Methods & ACC & ACC$_n$ & ACC$_l$ & ACC$_h$ & ACC$_t$\\

\hline
\rowcolor{gray!15} \multicolumn{6}{c}{\textit{Open-source LMMs}} \\
TextMonkey~\cite{liu2024textmonkey} $\diamond$ & 22.4 & 21.8 & 21.4 & 26.3 & 22.4 \\
mPLUG-DocOwl2~\cite{mplug-docowl2} $\diamond$ & 24.1 & 18.0 & 22.8 & 22.9 & 25.2 \\
LLaVA-Next-7B~\cite{llavaNext} & 17.9 & 18.9 & 22.7 & 24.6 & 20.4 \\
LLaVA-Next-13B~\cite{llavaNext} & 20.1 & 26.3 & 21.2 & 25.5 & 22.5 \\
LLaVA-OV-7B~\cite{llavaOV} & 31.3 & 26.7 & 26.3 & 31.6 & 30.7 \\
Qwen2-VL-7B~\cite{wang2024qwen2} & 39.7 & 37.5 & 29.0 & 39.0 & 38.4 \\
InternVL2.5-8B~\cite{internvl2_5} & 47.8 & 37.1 & 46.3 & 52.5 & 55.3 \\
InternVL2.5-26B~\cite{internvl2_5} & 53.6 & 42.7 & 41.4 & 60.1 & 56.2 \\
InternVL2.5-38B~\cite{internvl2_5} & 63.0 & 49.8 & 66.0 & 67.9 & 59.5 \\
\rowcolor{gray!15} \multicolumn{6}{c}{\textit{Proprietary LMMs}} \\
GPT-4o-20240806~\cite{openai2024gpt4o} & 68.1 & 49.3 & 66.2 & 71.3 & 66.1 \\
\hline
\hline
Average & 38.8 & 32.8 & 36.3 & 42.2 & 39.7 \\
\bottomrule[1pt]
\end{tabular}}
\caption{\textbf{Performances across various inference settings.} \textit{ACC}, \textit{ACC$_n$}, \textit{ACC$_l$}, \textit{ACC$_h$}, and \textit{ACC$_t$} correspond to the results obtained under the setting of answering with Chain-of-Thought (CoT), without CoT, cross-linguistic answering, answering with hint, and answering with task-specific instruction, respectively. For \textit{ACC$_l$}, \textit{ACC$_h$}, and \textit{ACC$_t$}, CoT is used by default.}
\label{tab:ablation}
\end{table}

\section{Experiments}
\subsection{Evaluation Protocols}
We use a general instruction set to guide LMMs to generate final answers in required format.
In cases where the answers deviate from the required format or they are excessively verbose, following previous works~\cite{lumathvista,mmlongbench}, we incorporate GPT-4o~\cite{openai2024gpt4o} as an answer extractor. Subsequently, the accuracy is computed to assess the model performance. For comprehensively evaluating LMMs' capabilities, we establish five distinct answering settings: 1) answering with Chain-of-Thought (CoT), 2) answering without CoT, 3) cross-linguistic answering, 4) answering with hint, and 5) answering with our customized task-specific instruction which directs the model to prioritize the textual content in the given image, as well as the objects, relationships, and constraints inherent to the question. \textit{\textbf{We show all instruction templates for clarity in the appendix.}}

\subsection{Baseline Models}
For open-source LMM, we select LLaVA series~\cite{llavaNext,llavaOV}, Qwen2-VL~\cite{wang2024qwen2}, InternVL2.5~\cite{internvl2_5}, TextMonkey~\cite{liu2024textmonkey}, and mPLUG-DocOwl2~\cite{mplug-docowl2}. In particular, TextMonkey and mPLUG-DocOwl2 are text-centric LMMs, orienting for OCR-related tasks. Other models are generic LMMs. 
In addition, the cutting-edge proprietary model, GPT-4o~\cite{openai2024gpt4o}, is also tested.

\subsection{Results and Analysis}
The main evaluation results are shown in Tab.~\ref{tab:main} and Tab.~\ref{tab:ablation}. In Tab.~\ref{tab:main}, the overall performance of each LMM is reported, accompanied by the breakdown performance on six reasoning types. In Tab.~\ref{tab:ablation}, the performance under five answering settings are demonstrated. We present some key observations and analysis on the main results as follows.

\noindent \textit{\textbf{\ding{182} LMMs desire enhancement on complex reasoning using OCR cues.}} 
In Tab.~\ref{tab:main}, with CoT reasoning, the open-source LMMs with 7B to 13B parameter size achieve less than 50\% accuracy on Reasoning-OCR. While InternVL2.5-8B~\cite{internvl2_5} can achieve 93.0\% on DocVQA and 84.8\% on ChartVQA, it only obtains 47.8\% performance on our Reasoning-OCR. The overall low performance of these LMMs could be attributed to the lack of OCR-related training samples that require reasoning in high complexity. Future works could explore constructing complex-reasoning-aware OCR training data in low-cost ways. 
Additionally, in terms of the model size, the larger LMM can achieve significantly better performance. InternVL2.5-38B achieves 63.0\%, surpassing its 8B variant by 15.2\%. 
To summarize, there is still huge potential for further improvement from the data-centric and reasoning strategy view.

\noindent \textit{\textbf{\ding{183} Text-centric LMMs lag far behind advanced generic LMMs on reasoning ability.}} 
As shown in Tab.~\ref{tab:main}, two text-centric LMMs, TextMonkey and mPLUG-DocOwl2, achieve overall 22.4\% and 24.1\% performance, respectively. In comparison, generic LMMs, Qwen2-VL-7B and InternVL2.5-8B, achieve 39.7\% and 47.8\% accuracy, respectively. There is a noticeable performance gap between text-centric and advanced generic LMMs. 
Benefiting from high-quality data scaling and modeling scheme evolution, generic LMMs demonstrate favorable reasoning capabilities using OCR cues. 
However, the text-centric models usually leverage a large amount of visual-text recognition and parsing data, as well as relatively simple VQA samples. The answers do not need concrete thinking steps, probably limiting the reasoning ability of text-centric LMMs. 

\noindent \textit{\textbf{\ding{184} LMMs fall short in decision reasoning.}} 
To explore the influence of reasoning type, we show the detailed performance on each reasoning type for each LMM in Tab.~\ref{tab:main}. 
Overall, we observe that LMMs fall short in decision reasoning. For instance, mPLUG-DocOwl2 and LLaVA-Next-7B cannot correctly answer any decision reasoning question. LLaVA-Next-13B, LLaVA-OV-7B, and InternVL2.5-8B only achieve 9.1\% accuracy. The best performing models are GPT-4o and InternVL2.5-38B, with 36.4\% accuracy. It indicates that LMMs are still hard to solve more complicated tasks in real-world, such as planning and decision making, which are important for embodied artificial intelligence and assisting people in their activity arrangements.

In addition to decision reasoning, LLaVA series and text-centric LMMs perform poorly on the mathematical reasoning with OCR cues. In comparison, recent advanced LMMs like Qwen2-VL and InternVL2.5 achieve significantly better mathematical performance.

Meanwhile, all LMMs are good at conditional reasoning. Moreover, InternVL2.5 series perform better on the data comparison analysis task compared to other reasoning types.

\begin{figure*}[t!]
\centering
\includegraphics[width=\linewidth]{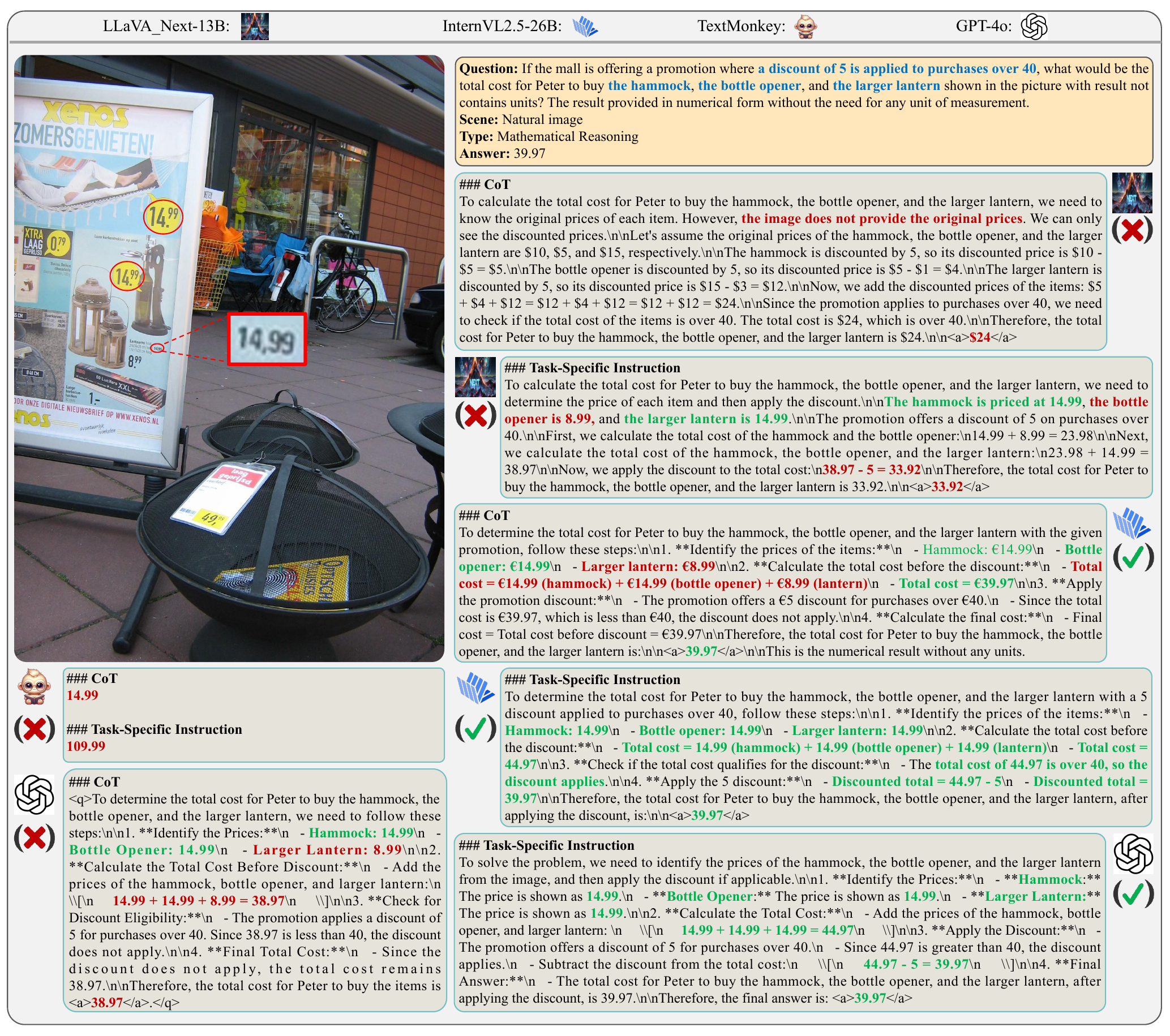} 
\caption{\textbf{Visualization of LMMs' responses on Reasoning-OCR under CoT and task-specific instruction settings.} Relevant textual cues in the image are highlighted with red circles or rectangles for clarity, which are not visible to LMMs. The key elements of the question are emphasized in \textbf{\darkblue{blue}}. In the responses generated by the LMMs, incorrect reasoning steps are marked in \textbf{\darkred{red}}, while correct reasoning steps are indicated in \textbf{\darkgreen{green}}. 
}
\vspace{-8pt}
\label{inf_example}
\end{figure*}

\noindent \textit{\textbf{\ding{185} CoT generally helps LMMs achieve better reasoning performance.}} 
By prompting LMMs to reasoning step-by-step, most LMMs obtain considerable performance gain. As shown in Tab.~\ref{tab:ablation}, the accuracy of LLaVA-OV-7B is increased by 4.6\% while InternVL2.5 series all achieve more than 10.0\% absolute improvement. It indicates that test-time scaling is useful for improving the complex reasoning performance of advanced LMMs. 
Some CoT answering samples are visualized in Fig.~\ref{inf_example}. With CoT, most LMMs break down the question and think step-by-step toward the final answering. However, TextMonkey only presents a final answer, which could be attributed to the fact that its training VQA samples are relatively easy and concise answers are enough.

\begin{figure*}[t!]
\centering
\includegraphics[width=\linewidth]{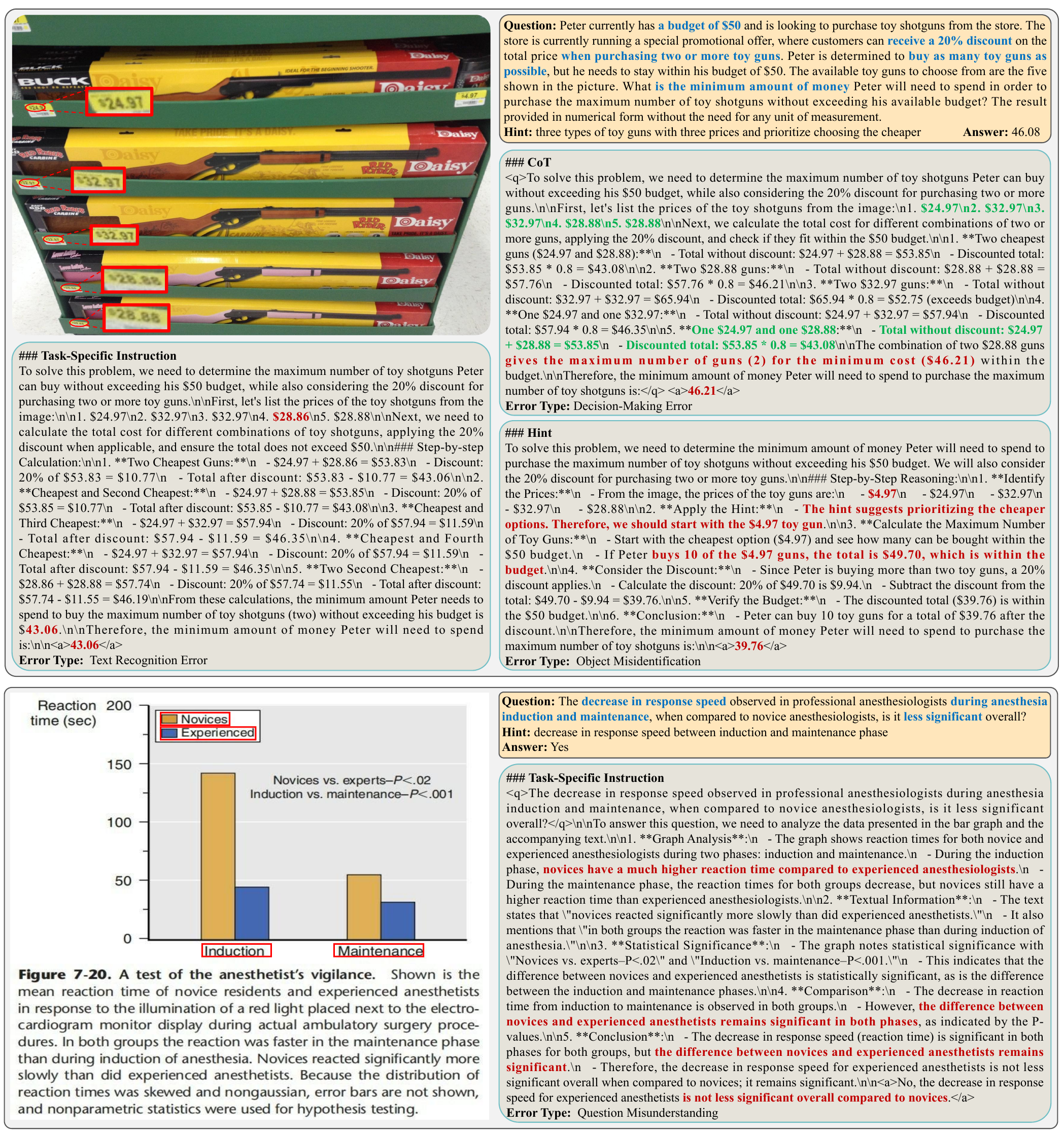} 
\caption{\textbf{Failure cases of GPT-4o on Reasoning-OCR.} Relevant textual cues in the images are highlighted with red circles or rectangles for clarity, which are not visible to LMMs. The key elements in questions are emphasized in \textbf{\darkblue{blue}}. In LMMs' responses, incorrect reasoning steps are marked in \textbf{\darkred{red}}.}
\label{false_example}
\vspace{-6pt}
\end{figure*}

\noindent \textit{\textbf{\ding{186} Providing the answer hint can improve the accuracy.}} 
In Tab.~\ref{tab:ablation}, providing the answer hint generally improves LMMs' reasoning performance. 
It indicates that LMMs could not find appropriate solution ideas based on the complex questions in some cases. The example for detailed hint content can be viewed in Fig.~\ref{false_example} . However, the answer hint may not be available in real-world application. Therefore, training the LMM to accurately analyze the problem, and form the inherent and correct problem-solving methodology may be promising.

\noindent \textit{\textbf{\ding{187} Designing task-specific instruction may be helpful.}}
The task-specific instruction can be referred to the appendix. As shown in Tab.~\ref{tab:ablation}, while instructing the model to pay close attention to the visual-text in image and explain the used text clues, some models obtain large performance gain, such as 7.5\% absolute improvement on InternVL2.5-8B. Therefore, more advanced prompt engineering could be explored to enhance the performance on specific difficult tasks.

\subsection{Error Source Analysis}
We summarize four major errors as follows and illustrate some failure cases in Fig.~\ref{false_example}.

\noindent \textbf{Object misidentification.} The model identifies or selects the wrong object in the image during reasoning.

\noindent \textbf{Text recognition error.} The model incorrectly recognizes the textual content within the image. In particular, the recognition on small and multi-oriented texts needs further improvement~\cite{yang2024cc}. Taking the text recognition error in Fig.~\ref{false_example} as an example, the small `28.88' text sequence is wrongly recognized as `28.86'.

\noindent \textbf{Decision-making error.} The model draws incorrect conclusions or makes flawed reasoning steps.

\noindent \textbf{Question misunderstanding.} When the question becomes complex, the model may fail to accurately comprehend the intent or requirements.

\section{Conclusion}
In this work, we introduce the Reasoning-OCR benchmark for assessing LMMs' complex reasoning capabilities using OCR cues while minimizing the impact of field-specialized knowledge. 
Reasoning-OCR covers diverse visual scenarios (including charts, product labels, documents, natural images, screen shot, and tokens) and contains six types of reasoning challenges, \ie, 1) data comparison analysis, 2) data statistical analysis, 3) mathematical reasoning, 4) conditional reasoning, 5) temporal reasoning, and 6) decision reasoning.  
By studying representative open-source and proprietary LMMs under five inference settings, we reveal that LMMs desire enhancement on complex reasoning using OCR cues. Comparing generic and text-centric LMMs, we observe that existing text-centric LMMs lag far behind recent advanced generic ones in reasoning ability. In terms of the specific reasoning type, existing LMMs typically fall short in decision reasoning but achieve promising progress in mathematical reasoning based on OCR cues.
We hope this work could inspire and facilitate future research on enhancing complex logical reasoning capabilities with OCR cues.

\appendix
\section*{Limitations}
Although our benchmark poses a significant challenge to existing LMMs and reveals insightful observations, the data could be sourced from a broader range to encompass more comprehensive scenarios.
Additionally, the complex logical questions require meticulous design, hindering the quick scaling of the question amount. With the growth of LMMs' reasoning ability, we anticipate an annotation procedure with a higher automation degree to scale up the benchmark volume in the future.

\section*{Ethical Statement}
The images in our benchmark are publicly available, and we ensure that the images do not contain any private information. This benchmark is only for research purpose.

\bibliographystyle{named}
\bibliography{ijcai25}

\begin{thebibliography}{}

\bibitem[\protect\citeauthoryear{Achiam \bgroup \em et al.\egroup }{2023}]{achiam2023gpt}
Josh Achiam, Steven Adler, Sandhini Agarwal, Lama Ahmad, Ilge Akkaya, et~al.
\newblock Gpt-4 technical report.
\newblock {\em arXiv preprint arXiv:2303.08774}, 2023.

\bibitem[\protect\citeauthoryear{Alayrac \bgroup \em et al.\egroup }{2022}]{alayrac2022flamingo}
Jean-Baptiste Alayrac, Jeff Donahue, Pauline Luc, Antoine Miech, et~al.
\newblock Flamingo: a visual language model for few-shot learning.
\newblock In {\em NeurIPS}, volume~35, pages 23716--23736, 2022.

\bibitem[\protect\citeauthoryear{Chen \bgroup \em et al.\egroup }{2021}]{chen2021geoqa}
Jiaqi Chen, Jianheng Tang, Jinghui Qin, Xiaodan Liang, Lingbo Liu, Eric Xing, and Liang Lin.
\newblock Geoqa: A geometric question answering benchmark towards multimodal numerical reasoning.
\newblock In {\em Findings of ACL-IJCNLP}, pages 513--523, 2021.

\bibitem[\protect\citeauthoryear{Chen \bgroup \em et al.\egroup }{2024a}]{chen2024onechart}
Jinyue Chen, Lingyu Kong, Haoran Wei, Chenglong Liu, Zheng Ge, et~al.
\newblock Onechart: Purify the chart structural extraction via one auxiliary token.
\newblock In {\em ACM MM}, pages 147--155, 2024.

\bibitem[\protect\citeauthoryear{Chen \bgroup \em et al.\egroup }{2024b}]{internvl2_5}
Zhe Chen, Weiyun Wang, Yue Cao, Yangzhou Liu, Zhangwei Gao, et~al.
\newblock Expanding performance boundaries of open-source multimodal models with model, data, and test-time scaling.
\newblock {\em arXiv preprint arXiv:2412.05271}, 2024.

\bibitem[\protect\citeauthoryear{Chen \bgroup \em et al.\egroup }{2024c}]{chen2024internvl}
Zhe Chen, Jiannan Wu, Wenhai Wang, Weijie Su, Guo Chen, et~al.
\newblock Internvl: Scaling up vision foundation models and aligning for generic visual-linguistic tasks.
\newblock In {\em CVPR}, pages 24185--24198, 2024.

\bibitem[\protect\citeauthoryear{Dai \bgroup \em et al.\egroup }{2023}]{dai2023instructblip}
Wenliang Dai, Junnan Li, D~Li, AMH Tiong, J~Zhao, W~Wang, B~Li, P~Fung, and S~Hoi.
\newblock Instructblip: Towards general-purpose vision-language models with instruction tuning. arxiv 2023.
\newblock {\em arXiv preprint arXiv:2305.06500}, 2, 2023.

\bibitem[\protect\citeauthoryear{He \bgroup \em et al.\egroup }{2024}]{he2024olympiadbench}
Chaoqun He, Renjie Luo, Yuzhuo Bai, Shengding Hu, Zhen~Leng Thai, et~al.
\newblock Olympiadbench: A challenging benchmark for promoting agi with olympiad-level bilingual multimodal scientific problems.
\newblock {\em arXiv preprint arXiv:2402.14008}, 2024.

\bibitem[\protect\citeauthoryear{Hu \bgroup \em et al.\egroup }{2024}]{mplug-docowl2}
Anwen Hu, Haiyang Xu, Liang Zhang, Jiabo Ye, Ming Yan, et~al.
\newblock mplug-docowl2: High-resolution compressing for ocr-free multi-page document understanding.
\newblock {\em arXiv preprint arXiv:2409.03420}, 2024.

\bibitem[\protect\citeauthoryear{Li \bgroup \em et al.\egroup }{2024a}]{li2024llava}
Bo~Li, Yuanhan Zhang, Dong Guo, Renrui Zhang, Feng Li, et~al.
\newblock Llava-onevision: Easy visual task transfer.
\newblock {\em arXiv preprint arXiv:2408.03326}, 2024.

\bibitem[\protect\citeauthoryear{Li \bgroup \em et al.\egroup }{2024b}]{llavaOV}
Bo~Li, Yuanhan Zhang, Dong Guo, Renrui Zhang, Feng Li, Hao Zhang, Kaichen Zhang, Peiyuan Zhang, Yanwei Li, Ziwei Liu, et~al.
\newblock Llava-onevision: Easy visual task transfer.
\newblock {\em arXiv preprint arXiv:2408.03326}, 2024.

\bibitem[\protect\citeauthoryear{Li \bgroup \em et al.\egroup }{2024c}]{li2024seed}
Bohao Li, Yuying Ge, Yi~Chen, Yixiao Ge, Ruimao Zhang, and Ying Shan.
\newblock Seed-bench-2-plus: Benchmarking multimodal large language models with text-rich visual comprehension.
\newblock {\em arXiv preprint arXiv:2404.16790}, 2024.

\bibitem[\protect\citeauthoryear{Li \bgroup \em et al.\egroup }{2024d}]{li2024monkey}
Zhang Li, Biao Yang, Qiang Liu, Zhiyin Ma, Shuo Zhang, et~al.
\newblock Monkey: Image resolution and text label are important things for large multi-modal models.
\newblock In {\em CVPR}, pages 26763--26773, 2024.

\bibitem[\protect\citeauthoryear{Liu \bgroup \em et al.\egroup }{2024a}]{liu2024focus}
Chenglong Liu, Haoran Wei, Jinyue Chen, Lingyu Kong, Zheng Ge, Zining Zhu, Liang Zhao, Jianjian Sun, Chunrui Han, and Xiangyu Zhang.
\newblock Focus anywhere for fine-grained multi-page document understanding.
\newblock {\em arXiv preprint arXiv:2405.14295}, 2024.

\bibitem[\protect\citeauthoryear{Liu \bgroup \em et al.\egroup }{2024b}]{liu2024mmc}
Fuxiao Liu, Xiaoyang Wang, Wenlin Yao, Jianshu Chen, Kaiqiang Song, et~al.
\newblock Mmc: Advancing multimodal chart understanding with large-scale instruction tuning.
\newblock In {\em NAACL}, pages 1287--1310, 2024.

\bibitem[\protect\citeauthoryear{Liu \bgroup \em et al.\egroup }{2024c}]{liu2024improved}
Haotian Liu, Chunyuan Li, Yuheng Li, and Yong~Jae Lee.
\newblock Improved baselines with visual instruction tuning.
\newblock In {\em CVPR}, pages 26296--26306, 2024.

\bibitem[\protect\citeauthoryear{Liu \bgroup \em et al.\egroup }{2024d}]{llavaNext}
Haotian Liu, Chunyuan Li, Yuheng Li, Bo~Li, Yuanhan Zhang, Sheng Shen, and Yong~Jae Lee.
\newblock Llava-next: Improved reasoning, ocr, and world knowledge, 2024.

\bibitem[\protect\citeauthoryear{Liu \bgroup \em et al.\egroup }{2024e}]{liu2024ocrbench}
Yuliang Liu, Zhang Li, Mingxin Huang, Biao Yang, Wenwen Yu, et~al.
\newblock Ocrbench: on the hidden mystery of ocr in large multimodal models.
\newblock {\em Science China Information Sciences}, 67(12):220102, 2024.

\bibitem[\protect\citeauthoryear{Liu \bgroup \em et al.\egroup }{2024f}]{liu2024textmonkey}
Yuliang Liu, Biao Yang, Qiang Liu, Zhang Li, et~al.
\newblock Textmonkey: An ocr-free large multimodal model for understanding document.
\newblock {\em arXiv preprint arXiv:2403.04473}, 2024.

\bibitem[\protect\citeauthoryear{Lu \bgroup \em et al.\egroup }{2024a}]{lumathvista}
Pan Lu, Hritik Bansal, Tony Xia, Jiacheng Liu, Chunyuan Li, et~al.
\newblock Mathvista: Evaluating mathematical reasoning of foundation models in visual contexts.
\newblock In {\em ICLR}, 2024.

\bibitem[\protect\citeauthoryear{Lu \bgroup \em et al.\egroup }{2024b}]{lu2024mathgenie}
Zimu Lu, Aojun Zhou, Houxing Ren, Ke~Wang, Weikang Shi, Junting Pan, Mingjie Zhan, and Hongsheng Li.
\newblock Mathgenie: Generating synthetic data with question back-translation for enhancing mathematical reasoning of llms.
\newblock {\em arXiv preprint arXiv:2402.16352}, 2024.

\bibitem[\protect\citeauthoryear{Lu \bgroup \em et al.\egroup }{2024c}]{lu2024mathcoder2}
Zimu Lu, Aojun Zhou, Ke~Wang, Houxing Ren, Weikang Shi, Junting Pan, Mingjie Zhan, and Hongsheng Li.
\newblock Mathcoder2: Better math reasoning from continued pretraining on model-translated mathematical code.
\newblock {\em arXiv preprint arXiv:2410.08196}, 2024.

\bibitem[\protect\citeauthoryear{Ma \bgroup \em et al.\egroup }{2024}]{mmlongbench}
Yubo Ma, Yuhang Zang, Liangyu Chen, Meiqi Chen, Yizhu Jiao, Xinze Li, Xinyuan Lu, Ziyu Liu, Yan Ma, Xiaoyi Dong, et~al.
\newblock Mmlongbench-doc: Benchmarking long-context document understanding with visualizations.
\newblock In {\em NeurIPS Datasets and Benchmarks Track}, 2024.

\bibitem[\protect\citeauthoryear{Masry \bgroup \em et al.\egroup }{2022}]{masry2022chartqa}
Ahmed Masry, Xuan~Long Do, Jia~Qing Tan, Shafiq Joty, and Enamul Hoque.
\newblock Chartqa: A benchmark for question answering about charts with visual and logical reasoning.
\newblock In {\em Findings of ACL}, pages 2263--2279, 2022.

\bibitem[\protect\citeauthoryear{Mathew \bgroup \em et al.\egroup }{2021}]{mathew2021docvqa}
Minesh Mathew, Dimosthenis Karatzas, and CV~Jawahar.
\newblock Docvqa: A dataset for vqa on document images.
\newblock In {\em WACV}, pages 2200--2209, 2021.

\bibitem[\protect\citeauthoryear{OpenAI}{2024}]{openai2024gpt4o}
OpenAI.
\newblock Gpt-4o system card, 2024.

\bibitem[\protect\citeauthoryear{Singh \bgroup \em et al.\egroup }{2019}]{singh2019towards}
Amanpreet Singh, Vivek Natarajan, Meet Shah, Yu~Jiang, Xinlei Chen, et~al.
\newblock Towards vqa models that can read.
\newblock In {\em CVPR}, pages 8317--8326, 2019.

\bibitem[\protect\citeauthoryear{Tang \bgroup \em et al.\egroup }{2024}]{tang2024mtvqa}
Jingqun Tang, Qi~Liu, Yongjie Ye, Jinghui Lu, Shu Wei, et~al.
\newblock Mtvqa: Benchmarking multilingual text-centric visual question answering.
\newblock {\em arXiv preprint arXiv:2405.11985}, 2024.

\bibitem[\protect\citeauthoryear{Wadhawan \bgroup \em et al.\egroup }{2024}]{wadhawancontextual}
Rohan Wadhawan, Hritik Bansal, Kai-Wei Chang, and Nanyun Peng.
\newblock Contextual: Evaluating context-sensitive text-rich visual reasoning in large multimodal models.
\newblock In {\em ICML}, 2024.

\bibitem[\protect\citeauthoryear{Wang \bgroup \em et al.\egroup }{2024a}]{wang2024measuring}
Ke~Wang, Junting Pan, Weikang Shi, Zimu Lu, Mingjie Zhan, and Hongsheng Li.
\newblock Measuring multimodal mathematical reasoning with math-vision dataset.
\newblock {\em arXiv preprint arXiv:2402.14804}, 2024.

\bibitem[\protect\citeauthoryear{Wang \bgroup \em et al.\egroup }{2024b}]{wang2024qwen2}
Peng Wang, Shuai Bai, Sinan Tan, Shijie Wang, Zhihao Fan, et~al.
\newblock Qwen2-vl: Enhancing vision-language model's perception of the world at any resolution.
\newblock {\em arXiv preprint arXiv:2409.12191}, 2024.

\bibitem[\protect\citeauthoryear{Wang \bgroup \em et al.\egroup }{2024c}]{wangcharxiv}
Zirui Wang, Mengzhou Xia, Luxi He, Howard Chen, Yitao Liu, et~al.
\newblock Charxiv: Charting gaps in realistic chart understanding in multimodal llms.
\newblock In {\em NeurIPS Datasets and Benchmarks Track}, 2024.

\bibitem[\protect\citeauthoryear{Wei \bgroup \em et al.\egroup }{2024}]{wei2024general}
Haoran Wei, Chenglong Liu, Jinyue Chen, Jia Wang, Lingyu Kong, et~al.
\newblock General ocr theory: Towards ocr-2.0 via a unified end-to-end model.
\newblock {\em arXiv preprint arXiv:2409.01704}, 2024.

\bibitem[\protect\citeauthoryear{Xia \bgroup \em et al.\egroup }{2024}]{xia2024chartx}
Renqiu Xia, Bo~Zhang, Hancheng Ye, Xiangchao Yan, Qi~Liu, et~al.
\newblock Chartx \& chartvlm: A versatile benchmark and foundation model for complicated chart reasoning.
\newblock {\em arXiv preprint arXiv:2402.12185}, 2024.

\bibitem[\protect\citeauthoryear{Xu \bgroup \em et al.\egroup }{2023}]{xu2023chartbench}
Zhengzhuo Xu, Sinan Du, Yiyan Qi, Chengjin Xu, Chun Yuan, and Jian Guo.
\newblock Chartbench: A benchmark for complex visual reasoning in charts.
\newblock {\em arXiv preprint arXiv:2312.15915}, 2023.

\bibitem[\protect\citeauthoryear{Yang \bgroup \em et al.\egroup }{2024a}]{yang2024if}
Ke~Yang, Jiateng Liu, John Wu, Chaoqi Yang, Yi~R Fung, Sha Li, Zixuan Huang, Xu~Cao, Xingyao Wang, Yiquan Wang, et~al.
\newblock If llm is the wizard, then code is the wand: A survey on how code empowers large language models to serve as intelligent agents.
\newblock {\em arXiv preprint arXiv:2401.00812}, 2024.

\bibitem[\protect\citeauthoryear{Yang \bgroup \em et al.\egroup }{2024b}]{yang2024cc}
Zhibo Yang, Jun Tang, Zhaohai Li, Pengfei Wang, Jianqiang Wan, et~al.
\newblock Cc-ocr: A comprehensive and challenging ocr benchmark for evaluating large multimodal models in literacy.
\newblock {\em arXiv preprint arXiv:2412.02210}, 2024.

\bibitem[\protect\citeauthoryear{Ye \bgroup \em et al.\egroup }{2023}]{ye2023ureader}
Jiabo Ye, Anwen Hu, Haiyang Xu, Qinghao Ye, et~al.
\newblock Ureader: Universal ocr-free visually-situated language understanding with multimodal large language model.
\newblock In {\em Findings of EMNLP}, pages 2841--2858, 2023.

\bibitem[\protect\citeauthoryear{Zhang \bgroup \em et al.\egroup }{2024a}]{zhang2024exploring}
Shuo Zhang, Biao Yang, Zhang Li, Zhiyin Ma, Yuliang Liu, and Xiang Bai.
\newblock Exploring the capabilities of large multimodal models on dense text.
\newblock In {\em ICDAR}, pages 281--298, 2024.

\bibitem[\protect\citeauthoryear{Zhang \bgroup \em et al.\egroup }{2024b}]{zhang2024unveiling}
Xinlu Zhang, Zhiyu~Zoey Chen, Xi~Ye, Xianjun Yang, Lichang Chen, William~Yang Wang, and Linda~Ruth Petzold.
\newblock Unveiling the impact of coding data instruction fine-tuning on large language models reasoning.
\newblock {\em arXiv preprint arXiv:2405.20535}, 2024.

\bibitem[\protect\citeauthoryear{Zhang \bgroup \em et al.\egroup }{2025}]{zhang2025mathverse}
Renrui Zhang, Dongzhi Jiang, Yichi Zhang, Haokun Lin, Ziyu Guo, et~al.
\newblock Mathverse: Does your multi-modal llm truly see the diagrams in visual math problems?
\newblock In {\em ECCV}, pages 169--186, 2025.

\end{thebibliography}

\newpage

\begin{figure*}[t]
\centering
\includegraphics[width=\linewidth]{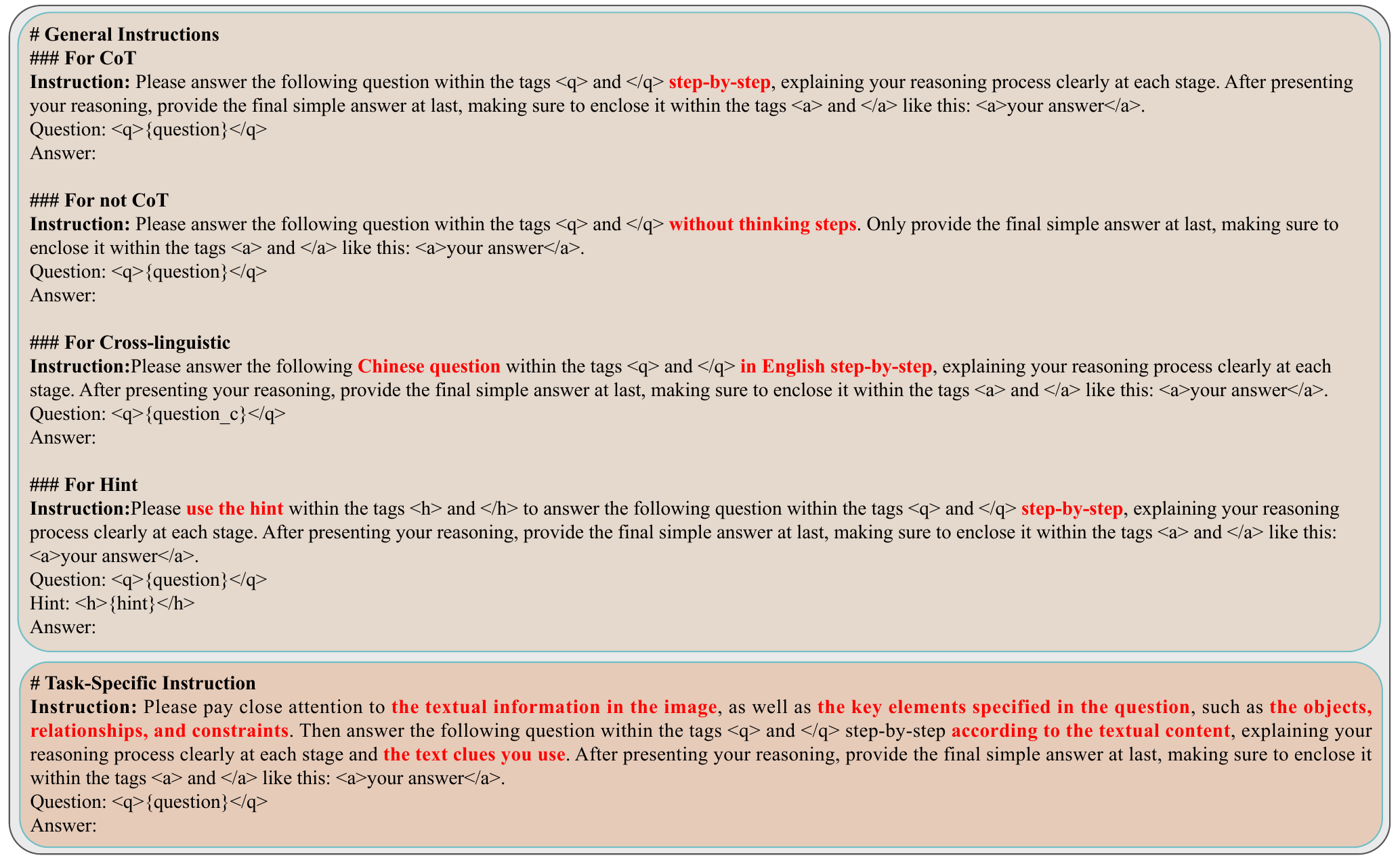} 
\caption{\textbf{Full instructions under different answering settings.}}
\label{fig: instruct}
\end{figure*}

\section{Instruction Templates}
For clarity, the full instructions under five answering settings are shown in Fig.~\ref{fig: instruct}.

\section{More Examples}
\begin{figure*}[t!]
\centering
\includegraphics[width=\linewidth]{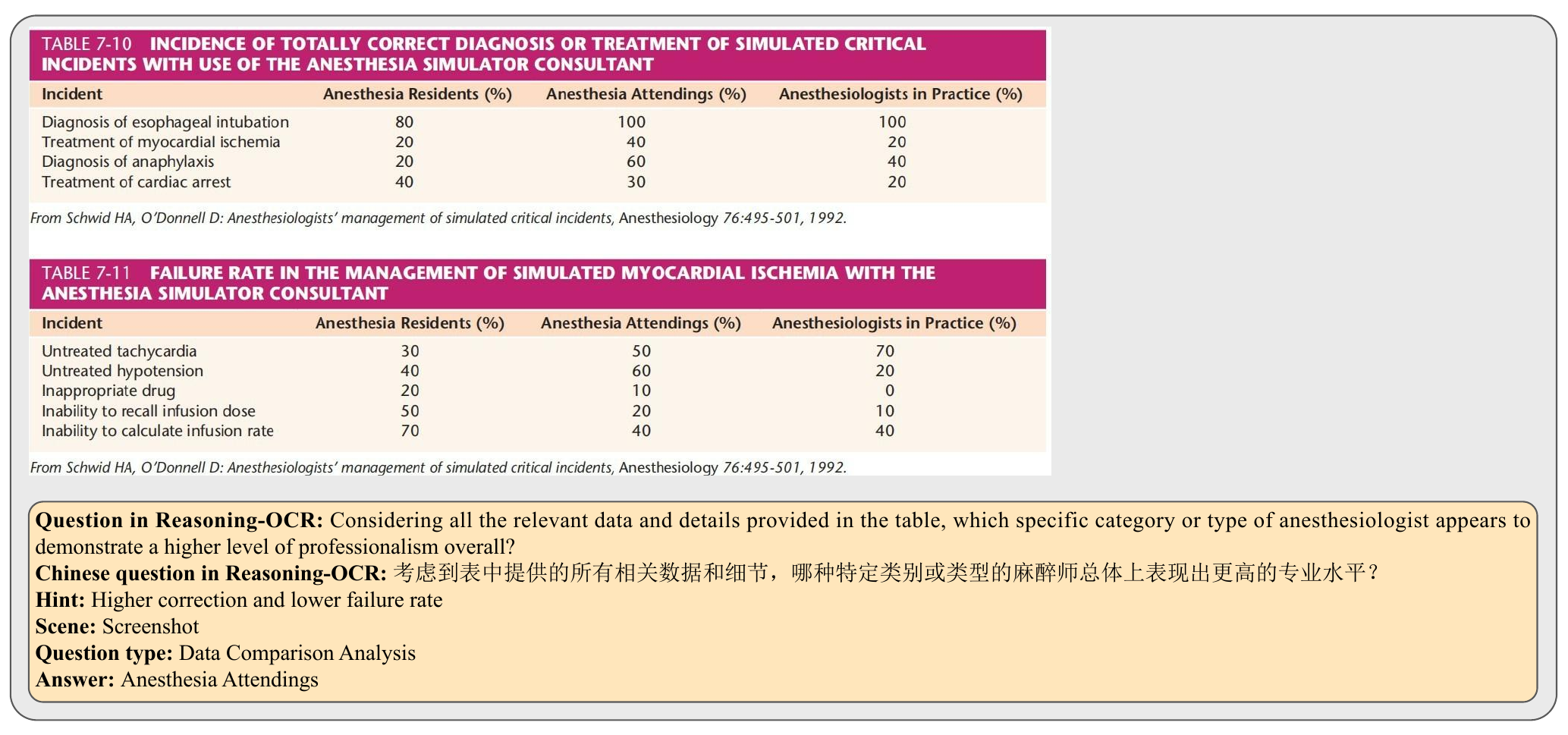} 
\caption{\textbf{An example for data comparison analysis.}}
\label{fig: dc}
\end{figure*}

\begin{figure*}[t!]
\centering
\includegraphics[width=\linewidth]{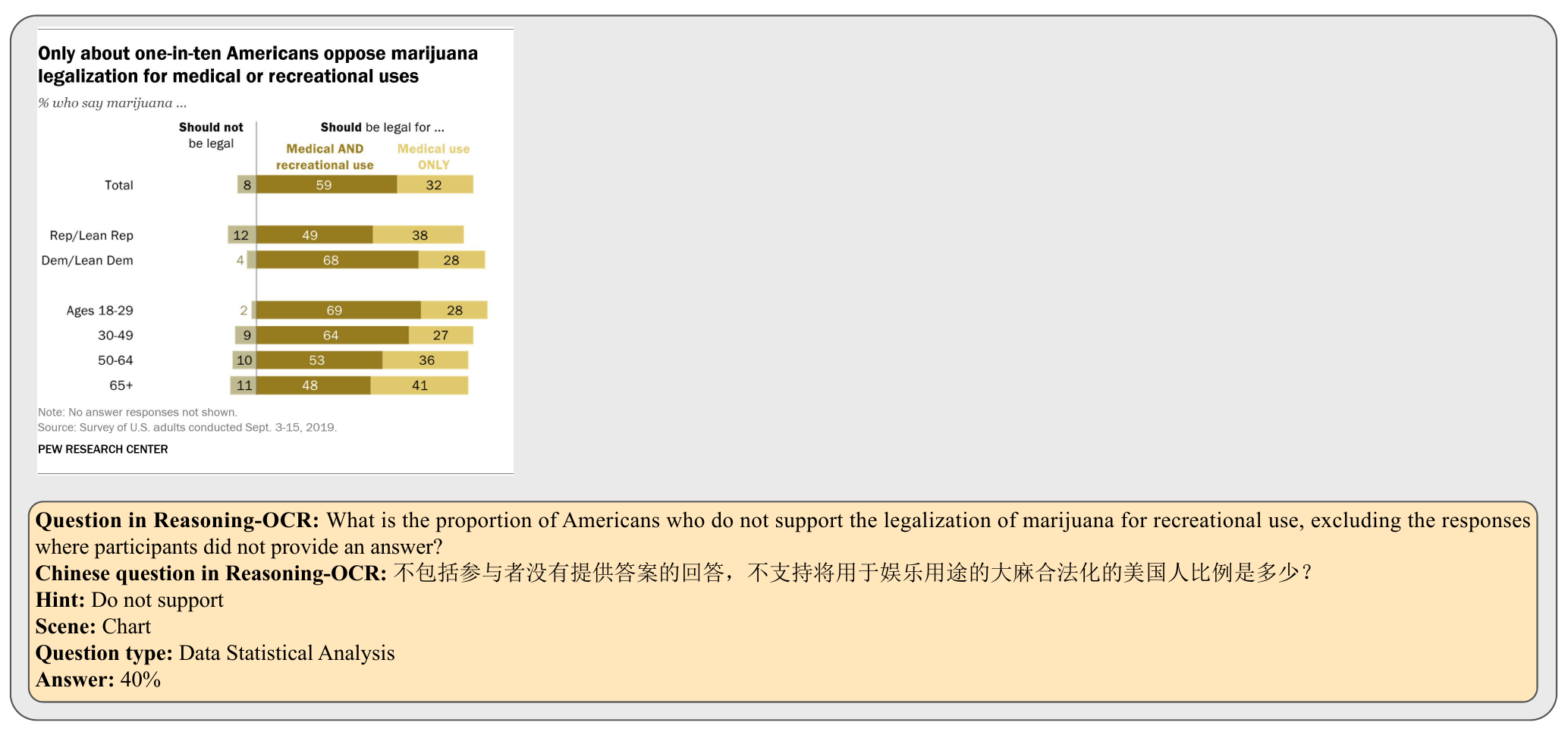} 
\caption{\textbf{An example for data statistical analysis.}}
\label{fig: ds}
\end{figure*}

\begin{figure*}[t!]
\centering
\includegraphics[width=\linewidth]{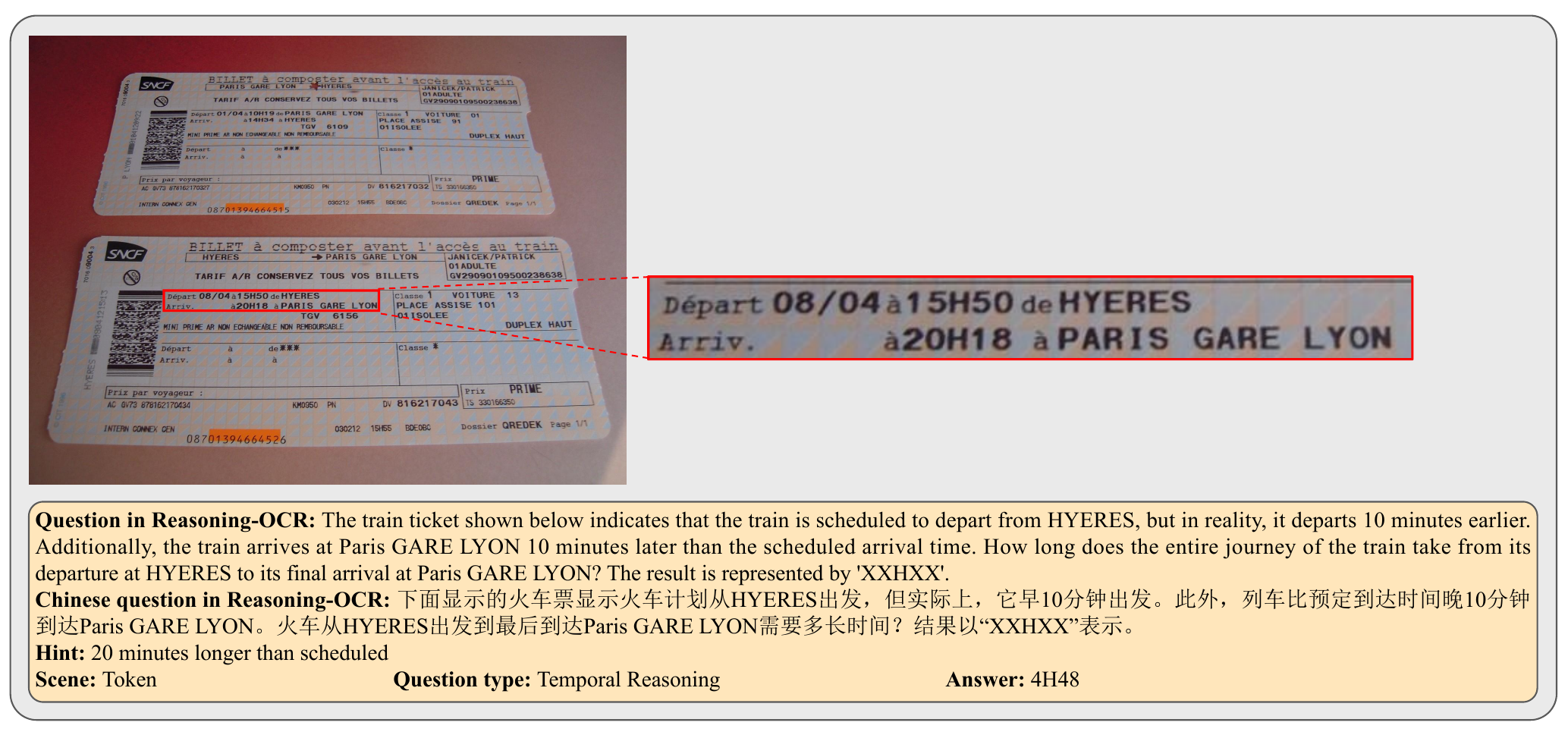} 
\caption{\textbf{An example for mathematical reasoning.} Relevant textual cues in the image are highlighted with red circles or rectangles for clarity, which are not visible to LMMs.}
\label{fig: mr}
\end{figure*}

\begin{figure*}[t!]
\centering
\includegraphics[width=\linewidth]{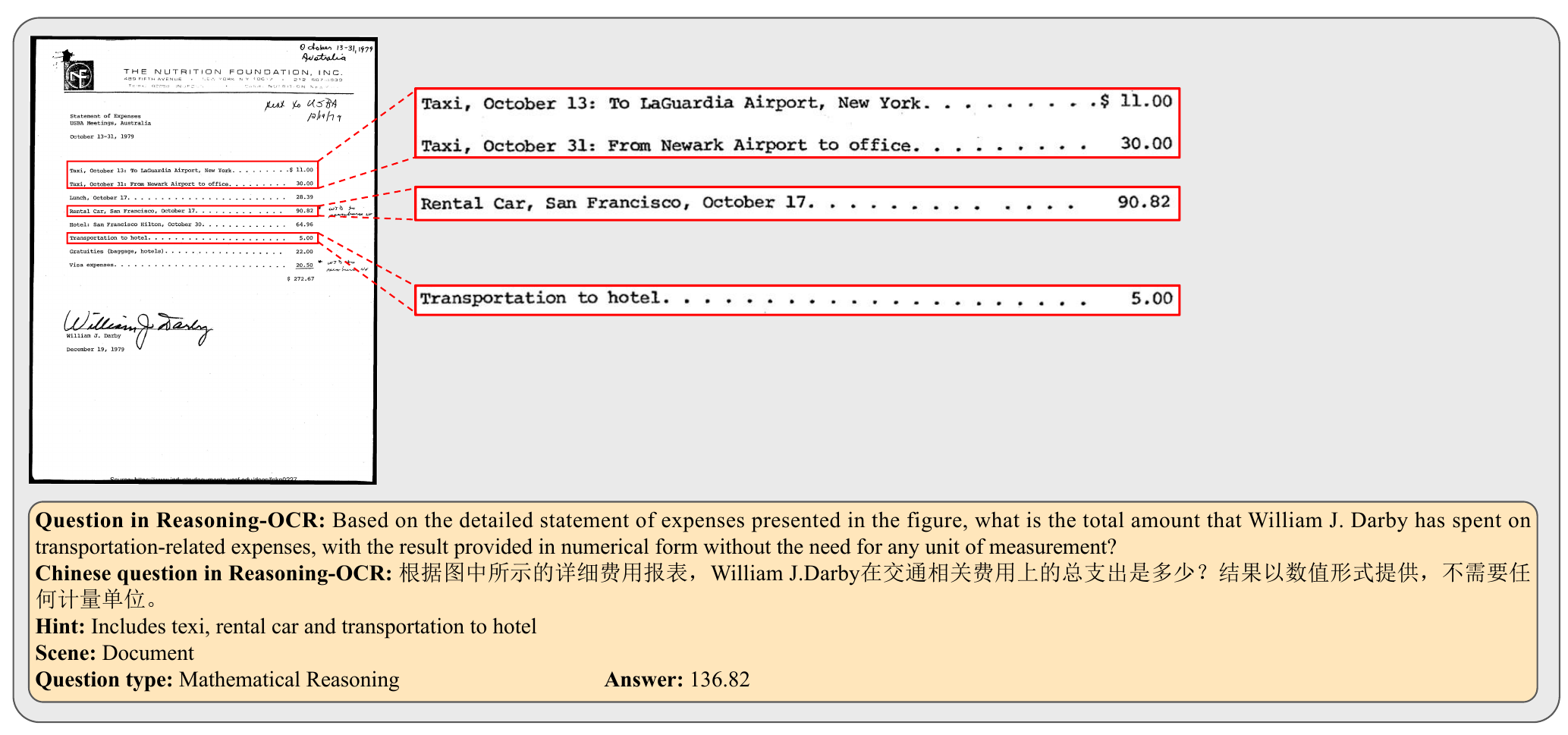} 
\caption{\textbf{An example for temporal reasoning.} Relevant textual cues in the image are highlighted with red circles or rectangles for clarity, which are not visible to LMMs.}
\label{fig: tr}
\end{figure*}

\begin{figure*}[t!]
\centering
\includegraphics[width=\linewidth]{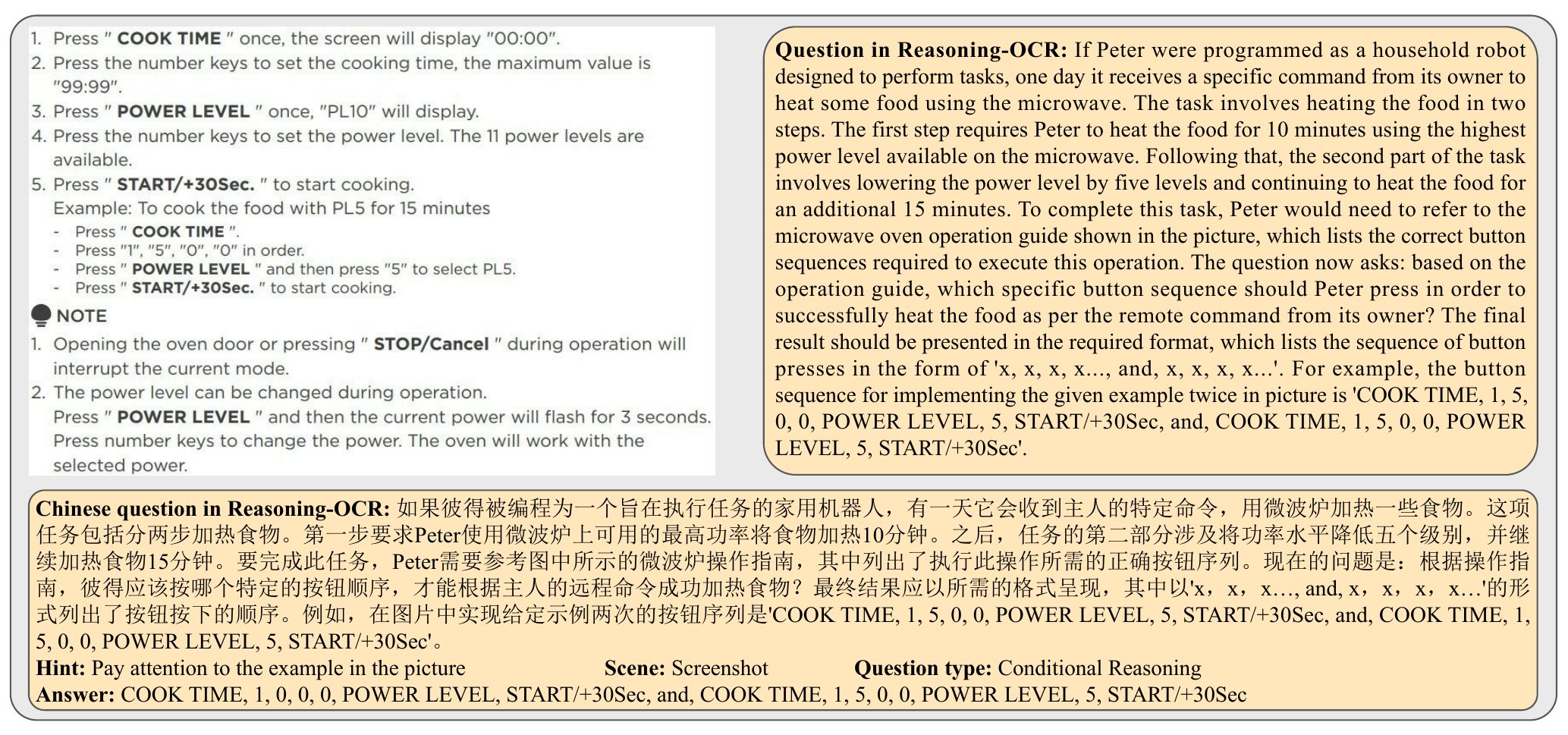} 
\caption{\textbf{An example for conditional reasoning.}}
\label{fig: cr}
\end{figure*}

\begin{figure*}[t!]
\centering
\includegraphics[width=\linewidth]{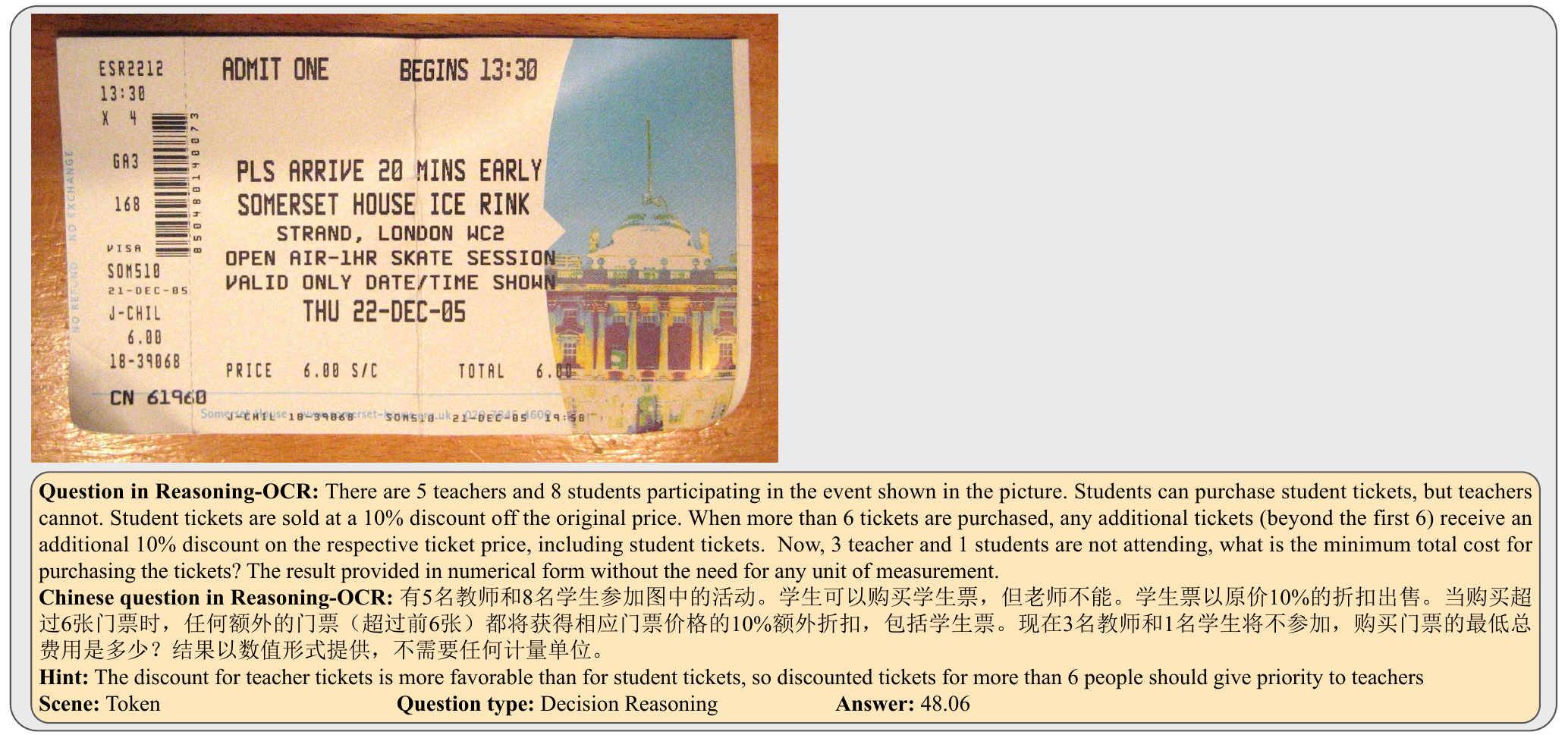} 
\caption{\textbf{An example for decision reasoning.}}
\label{fig: dr}
\end{figure*}

More examples for six reasoning types are shown in Fig.~\ref{fig: dc} to Fig.~\ref{fig: dr}.

\section{Attempts of GPT in Question Generation}
\begin{figure*}
    \centering
    \includegraphics[width=0.95\linewidth]{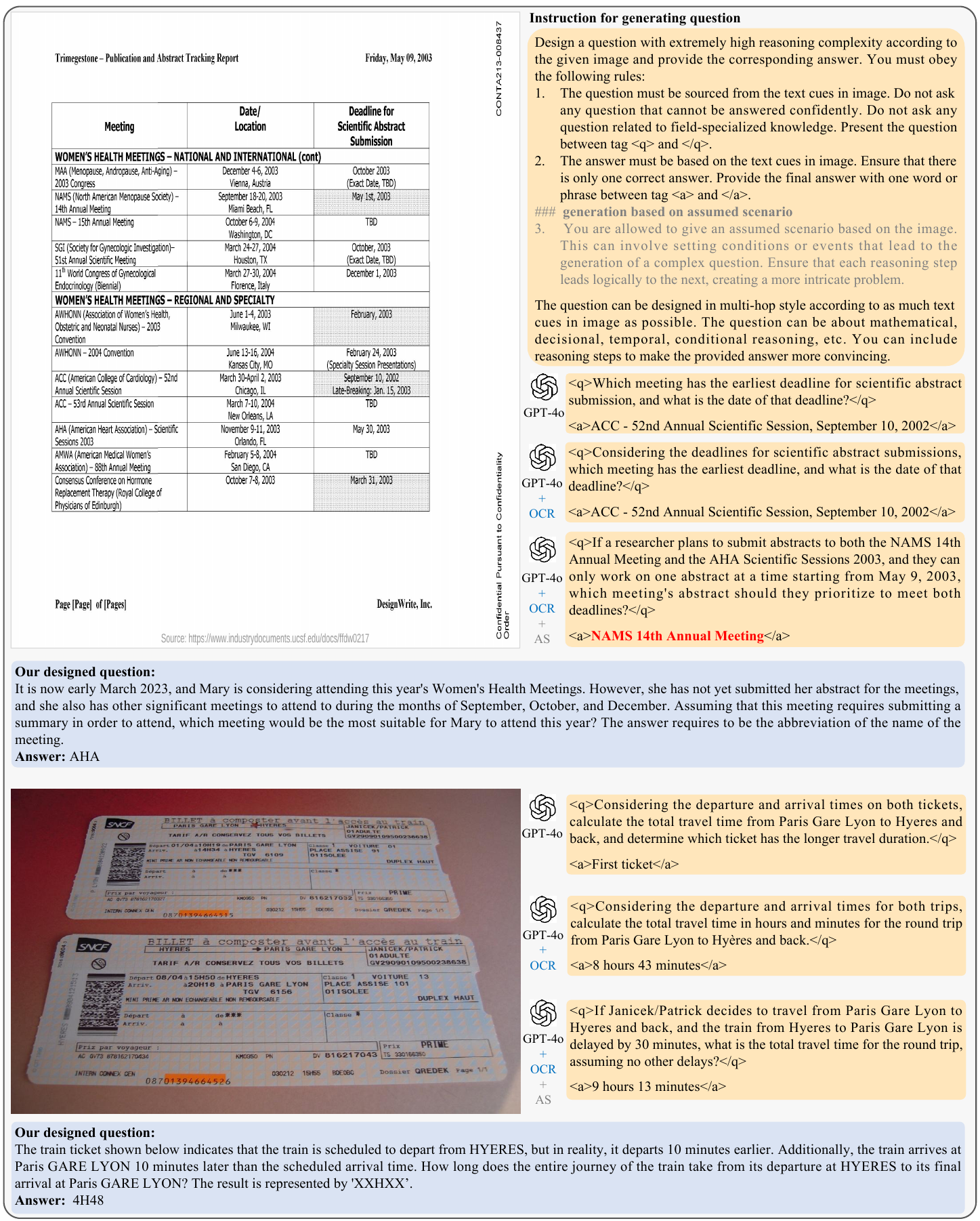}
    \caption{Examples of generating complex reasoning questions based on OCR cues in images using GPT-4o. \textcolor{blue}{\textbf{OCR}} denotes the use of an OCR model to extract textual information from images for supporting question generation. \textcolor{gray}{\textbf{AS}} indicates to enabling GPT-4o to assume scenarios and incorporate additional conditions, consequently increasing the complexity of the generated questions. Text in \textcolor{red}{red} means the error answer provided by GPT-4o.}
    \label{fig:gpt_gen}
\end{figure*}

\quad We investigate the use of GPT-4o~\cite{openai2024gpt4o} to generate complex multi-hop logical reasoning questions derived from textual cues within images, guided by different instructions as illustrated in Fig.~\ref{fig:gpt_gen}. The results indicate that while integrating the OCR model and scene assumptions has substantially improved GPT-4o's question-generation capabilities, it still significantly lags behind human-designed questions in terms of complexity. Moreover, it might provide the error answer as shown in the figure. This underscores the inherent challenge of scaling complex logical reasoning question-answering datasets with current large language models, which remains an open research challenge. 

Analogous to the ``chicken and egg" dilemma, both question generation and answering are manifestations of underlying logical reasoning abilities. However, designing intricate, multi-hop reasoning questions probably demands a higher level of abstract thinking and reasoning. Consequently, generating complex reasoning questions may be inherently more difficult than answering them.
In light of these observations, future research is warranted to further explore ways to enhance the capabilities of these models. Recent studies~\cite{zhang2024unveiling,lu2024mathgenie,yang2024if,lu2024mathcoder2} suggest that improving model performance in specific domains, such as mathematics and coding, can have a broader positive impact on their general reasoning capabilities. Such findings present a promising direction for advancing the complexity of text-based logical reasoning question generation.


\section{Datasheet for Reasoning-OCR
\label{sec_datasheet}}
\subsection{Motivation} 
\noindent 1. \textit{For what purpose was the dataset created?} \vspace{0.5\baselineskip}\\
\noindent\textbf{A1:} 
Reasoning-OCR is created to evaluate the complex logical reasoning abilities of large multimodal models (LMMs), particularly their multi-hop reasoning and decision-making capabilities, which are not fully assessed by existing visual-text benchmarks. It provides a novel benchmark to evaluate LMMs' reasoning performance using textual cues within images.
\vspace{0.5\baselineskip}

\noindent 2. \textit{Who created the dataset (e.g., which team, research group) and on behalf of which entity (e.g., company, institution, organization)? } \vspace{0.5\baselineskip}\\
\noindent\textbf{A2:} The dataset was created by the authors.\vspace{0.5\baselineskip}

\noindent 3. \textit{Who funded the creation of the dataset? }\vspace{0.5\baselineskip}\\
\noindent\textbf{A3:} The dataset creation was funded by the affiliations of the authors involved in this work.

\subsection{Composition}
\noindent 1. \textit{What do the instances that comprise the dataset represent (e.g., documents, photos, people, countries)? Are there multiple types of instances(e.g., movies, users, and ratings; people and interactions between them; nodes and edges)? Please provide a description. }\vspace{0.5\baselineskip}\\
\noindent\textbf{A1:}  
This dataset is composed of images from ChartQA, DocVQA, DT-VQA, and various websites, featuring charts, product labels, documents, natural images, screenshots, and tokens (such as paper currency and tickets).

\noindent 2. \textit{How many instances are there in total (of each type, if appropriate)?}\vspace{0.5\baselineskip}\\
\noindent\textbf{A2:} 
Reasoning-OCR consists of 140 curated images, categorized into 62 charts, 23 product labels, 22 documents, 18 natural images, 9 screenshots, and 6 tokens.
\vspace{0.5\baselineskip}

\noindent 3. \textit{Does the dataset contain all possible instances or is it a sample (not necessarily random) of instances from a larger set? }\vspace{0.5\baselineskip}\\
\noindent\textbf{A3:} 
No. The Reasoning-OCR dataset consists of four parts: 60 images (42.9\%) from ChartQA, 50 images (35.7\%) from DT-VQA, 20 images (14.3\%) from DocVQA, and 10 images (7.1\%) sourced from the web.
\vspace{0.5\baselineskip}

\noindent 4. \textit{Is there a label or target associated with each instance? If so, please provide a description.}\vspace{0.5\baselineskip}\\
\noindent\textbf{A4:} 
Yes. Each instance in the dataset is annotated with the following attributes: 1) \textit{img}: image name; 2) \textit{q\_id}: question index; 3) \textit{question}: question content; 4) \textit{question\_c}: Chinese version of the question; 5) \textit{answer}: concise answer; 6) \textit{hint}: a helpful hint; 7) \textit{datasource}: image source; 8) \textit{scene}: scene category; 9) \textit{type}: question type.
\vspace{0.5\baselineskip}

\noindent 5. \textit{Is any information missing from individual instances? If so, please provide a description, explaining why this information is missing (e.g., because it was unavailable). This does not include intentionally removed information, but might include, e.g., redacted text.}\vspace{0.5\baselineskip}\\
\noindent\textbf{A5:} No. \vspace{0.5\baselineskip}

\noindent 6. \textit{Are relationships between individual instances made explicit (e.g., users’ movie ratings, social network links)? If so, please describe how these relationships are drawn.}\vspace{0.5\baselineskip}\\
\noindent\textbf{A6:} Yes. The relationships between different instances are shown in filenames. \vspace{0.5\baselineskip}

\noindent 7. \textit{Are there recommended data splits (e.g., training, development/validation, testing)? If so, please provide a description of these splits, explaining the rationale behind them.}\vspace{0.5\baselineskip}\\
\noindent\textbf{A7:} No. The entire dataset is specifically designed to evaluate the reasoning capabilities of LMMs; as such, it consists solely of a test set. \vspace{0.5\baselineskip}

\noindent 8. \textit{Are there any errors, sources of noise, or redundancies in the dataset? If so, please provide a description.}\vspace{0.5\baselineskip}\\
\noindent\textbf{A8:} No. \vspace{0.5\baselineskip}

\noindent 9. \textit{Is the dataset self-contained, or does it link to or otherwise rely on external resources (e.g., websites, tweets, other datasets)?}\vspace{0.5\baselineskip}\\
\noindent\textbf{A9:} The dataset is self-contained. \vspace{0.5\baselineskip}

\noindent 10. \textit{Does the dataset contain data that might be considered confidential (e.g., data that is protected by legal privilege or by doctor-patient confidentiality, data that includes the content of individuals non-public communications)? If so, please provide a description.}\vspace{0.5\baselineskip}\\
\noindent\textbf{A10:} No. \vspace{0.5\baselineskip}

\noindent 11. \textit{Does the dataset contain data that, if viewed directly, might be offensive, insulting, threatening, or might otherwise cause anxiety? If so, please describe why.}\vspace{0.5\baselineskip}\\
\noindent\textbf{A11:} No. \vspace{0.5\baselineskip}

\subsection{Collection process} 
\noindent 1. \textit{How was the data associated with each instance acquired? }\vspace{0.5\baselineskip}\\
\noindent\textbf{A1:} Please refer to the details listed in the main text Sec.~3. \vspace{0.5\baselineskip}

\noindent 2. \textit{What mechanisms or procedures were used to collect the data (e.g., hardware apparatus or sensor, manual human curation, software program, software API)? How were these mechanisms or procedures validated?} \vspace{0.5\baselineskip}\\
\noindent\textbf{A2:} Please refer to the details listed in the main text Sec.~3.\vspace{0.5\baselineskip}

\noindent 3. \textit{If the dataset is a sample from a larger set, what was the sampling strategy (e.g., deterministic, probabilistic with specific sampling probabilities)?} \vspace{0.5\baselineskip}\\
\noindent\textbf{A3:}
The images suitable for generating multi-hop inference problems are sampled from the data sources.
\vspace{0.5\baselineskip}

\noindent 4. \textit{Who was involved in the data collection process (e.g., students, crowdworkers, contractors) and how were they compensated (e.g., how much were crowdworkers paid)} \vspace{0.5\baselineskip}\\
\noindent\textbf{A4:} The data was collected and verified by the authors.\vspace{0.5\baselineskip}

\subsection{Preprocessing/cleaning/labeling} 
\noindent 1. \textit{Was any preprocessing/cleaning/labeling of the data done (e.g., discretization or bucketing, tokenization, part-of-speech tagging, SIFT feature extraction, removal of instances, processing 5 of missing values)? If so, please provide a description. If not, you may skip the remainder of the questions in this section.}\vspace{0.5\baselineskip}\\
\noindent\textbf{A1:} Yes. We select images from the data sources that are conducive to generating multi-hop reasoning questions for our dataset. \vspace{0.5\baselineskip}

\noindent 2. \textit{Was the “raw” data saved in addition to the preprocessed/cleaned/labeled data (e.g., to support unanticipated future uses)? If so, please provide a link or other access point to the “raw” data.}\vspace{0.5\baselineskip}\\
\noindent\textbf{A2:} No.\vspace{0.5\baselineskip}

\noindent 3. \textit{Is the software used to preprocess/clean/label the instances available? } \vspace{0.5\baselineskip}\\
\noindent\textbf{A3:} Yes. We use 'Sublime Text' to label the data.\vspace{0.5\baselineskip}

\subsection{Uses} 
\noindent 1. \textit{Has the dataset been used for any tasks already? If so, please provide a description.}\vspace{0.5\baselineskip}\\
\noindent\textbf{A1:} No.
\vspace{0.5\baselineskip}

\noindent 2. \textit{Is there a repository that links to any or all papers or systems that use the dataset? If so, please provide a link or other access point.
}\vspace{0.5\baselineskip}\\
\noindent\textbf{A2:} N/A.\vspace{0.5\baselineskip}

\noindent 3. \textit{What (other) tasks could the dataset be used for?
} \vspace{0.5\baselineskip}\\
\noindent\textbf{A3:} 
No.\vspace{0.5\baselineskip}

\noindent 4. \textit{Is there anything about the composition of the dataset or the way it was collected and preprocessed/cleaned/labeled that might impact future uses? For example, is there anything that a future user might need to know to avoid uses that could result in unfair treatment of individuals or groups (e.g., stereotyping, quality of service issues) or other undesirable harms (e.g., financial harms, legal risks) If so, please provide a description. Is there anything a future user could do to mitigate these undesirable harms?
} \vspace{0.5\baselineskip}\\
\noindent\textbf{A4:} No.
\vspace{0.5\baselineskip}

\noindent 5. \textit{Are there tasks for which the dataset should not be used? If so, please provide a description.
} \vspace{0.5\baselineskip}\\
\noindent\textbf{A5:} No.
\vspace{0.5\baselineskip}

\subsection{Distribution} 
\noindent 1. \textit{Will the dataset be distributed to third parties outside of the entity (e.g., company, institution, organization) on behalf of which the dataset was created?
}\vspace{0.5\baselineskip}\\
\noindent\textbf{A1:} Yes. The dataset will be publicly available.
\vspace{0.5\baselineskip}

\noindent 2. \textit{How will the dataset be distributed (e.g., tarball on website, API, GitHub)?}\vspace{0.5\baselineskip}\\
\noindent\textbf{A2:} It will be publicly available on the GitHub.
\vspace{0.5\baselineskip}

\noindent 3. \textit{When will the dataset be distributed?
}\vspace{0.5\baselineskip}\\
\noindent\textbf{A3:} The dataset will be distributed once the paper is accepted after peer review.
\vspace{0.5\baselineskip}

\noindent 4. \textit{Will the dataset be distributed under a copyright or other intellectual property (IP) license, and/or under applicable terms of use (ToU)? If so, please describe this license and/or ToU, and provide a link or other access point to, or otherwise reproduce, any relevant licensing terms or ToU, as well as any fees associated with these restrictions.
}\vspace{0.5\baselineskip}\\
\noindent\textbf{A4:} It will be distributed under the \href{https://creativecommons.org/licenses/by-nc-sa/4.0/}{Creative Commons Attribution-NonCommercial-ShareAlike 4.0 License}.
\vspace{0.5\baselineskip}

\noindent 5. \textit{Have any third parties imposed IP-based or other restrictions on the data associated with the instances? If so, please describe these restrictions, and provide a link or other access point to, or otherwise reproduce, any relevant licensing terms, as well as any fees associated with these restrictions. 
}\vspace{0.5\baselineskip}\\
\noindent\textbf{A5:} No.
\vspace{0.5\baselineskip}

\noindent 6. \textit{Do any export controls or other regulatory restrictions apply to the dataset or to individual instances? If so, please describe these restrictions, and provide a link or other access point to, or otherwise reproduce, any supporting documentation.
}\vspace{0.5\baselineskip}\\
\noindent\textbf{A6:} No.
\vspace{0.5\baselineskip}

\subsection{Maintenance} 
\noindent 1. \textit{Who will be supporting/hosting/maintaining the dataset? }\vspace{0.5\baselineskip}\\
\noindent\textbf{A1:} The authors.
\vspace{0.5\baselineskip}

\noindent 2. \textit{How can the owner/curator/manager of the dataset be contacted (e.g., email address)? 
}\vspace{0.5\baselineskip}\\
\noindent\textbf{A2:} They can be contacted via the email address provided in the paper upon its acceptance following the peer review process.
\vspace{0.5\baselineskip}

\noindent 3. \textit{Is there an erratum? If so, please provide a link or other access point. 
}\vspace{0.5\baselineskip}\\
\noindent\textbf{A3:} No. 
\vspace{0.5\baselineskip}

\noindent 4. \textit{Will the dataset be updated (e.g., to correct labeling errors, add new instances, delete instances)? If so, please describe how often, by whom, and how updates will be communicated to users (e.g., mailing list, GitHub)? 
}\vspace{0.5\baselineskip}\\
\noindent\textbf{A4:} No. 
\vspace{0.5\baselineskip}

\noindent 5. \textit{Will older versions of the dataset continue to be supported/hosted/maintained? If so, please describe how. If not, please describe how its obsolescence will be communicated to users. 
}\vspace{0.5\baselineskip}\\
\noindent\textbf{A5:} N/A. 
\vspace{0.5\baselineskip}

\noindent 6. \textit{If others want to extend/augment/build on/contribute to the dataset, is there a mechanism for them to do so? If so, please provide a description. Will these contributions be validated/verified? If so, please describe how. If not, why not? Is there a process for communicating/distributing these contributions to other users? If so, please provide a description. 
}\vspace{0.5\baselineskip}\\
\noindent\textbf{A6:} N/A. 
\vspace{0.5\baselineskip}

\end{document}